\documentclass[10pt,twocolumn,twoside]{IEEEtran}

\usepackage{amsmath,amssymb,amsfonts}
\usepackage{amsthm}
\usepackage{mathrsfs}

\usepackage{graphicx}
\graphicspath{{./IMAGES/}} 
\usepackage[table]{xcolor}
\usepackage{array}
\usepackage{multirow}
\usepackage{booktabs}
\usepackage{makecell}

\usepackage{stfloats}
\usepackage{balance}

\usepackage{pifont}
\usepackage{textcomp}
\usepackage{url}
\usepackage{verbatim}
\usepackage{tikz}

\usepackage[ruled,lined,linesnumbered,shortend]{algorithm2e}
\DontPrintSemicolon
\SetAlgoLined
\SetAlgoShortEnd
\SetKwInput{KwInput}{Input}
\SetKwInput{KwOutput}{Output}
\SetKwInput{KwInit}{Init}
\SetKwFor{For}{for}{do}{end}
\SetKwFor{ForEach}{for each}{do}{end}
\definecolor{stagered}{HTML}{C62828}

\usepackage{hyperref}
\hypersetup{hidelinks,
  pdftitle={Continual Learning with Vision-Language Models via Semantic-Geometry Preservation},
  pdfauthor={Chiyuan He, Zihuan Qiu, Fanman Meng, Runtong Zhang, Linfeng Xu, Qingbo Wu, and Hongliang Li},
  pdfsubject={IEEE TCSVT manuscript},
  pdfkeywords={continual learning, vision-language model, semantic geometry, knowledge distillation}
}

\def\BibTeX{{\rm B\kern-.05em{\sc i\kern-.025em b}\kern-.08em
    T\kern-.1667em\lower.7ex\hbox{E}\kern-.125emX}}

\makeatletter
\renewcommand{\maketag@@@}[1]{\hbox{\m@th\normalsize\normalfont#1}}%
\makeatother

\begin{document}

%\title{VLM-Based Continual Learning with Semantic Geometry Preservation}
\title{Continual Learning with Vision-Language Models via Semantic-Geometry Preservation}
%\author{Chiyuan He, Zihuan Qiu, Fanman Meng, \textit{Member, IEEE}, Linfeng Xu, \textit{Member, IEEE}, \\ Qingbo Wu, \textit{Member, IEEE}, Hongliang Li, \textit{Senior Member, IEEE}
\author{Chiyuan He, Zihuan Qiu, Fanman Meng, \textit{Member, IEEE}, Runtong Zhang, Linfeng Xu, \textit{Member, IEEE}, \\ Qingbo Wu, \textit{Member, IEEE}, Hongliang Li, \textit{Senior Member, IEEE}
\thanks{Copyright \textcopyright{} 2026 IEEE. Personal use of this material is permitted. However, permission to use this material for any other purposes must be obtained from the IEEE by sending an email to pubs-permissions@ieee.org.}
 \thanks{Corresponding author: Fanman Meng. The authors are with the School of Information and Communication Engineering, University of Electronic Science and Technology of China, Chengdu 611731, China (e-mail: \{cyhe;zihuanqiu;rtzhang\}@std.uestc.edu.cn; \{fmmeng;lfxu;qbwu;hlli\}@uestc.edu.cn.
} 
\thanks{This work was supported in part by the National Natural Science Foundation of China under Grant (62271119, 62577016, U23A20286), the Key Research and Development Project of Hainan Province under Grant ZDYF2024(LALH)003, and the Natural Science Foundation of Sichuan
Province (No. 2025ZNSFSC0475).}
}

\maketitle

\begin{abstract}

Continual learning of pretrained vision-language models (VLMs) is prone to catastrophic forgetting, yet existing approaches adapt to new tasks without explicitly preserving the cross-modal semantic geometry inherited from pretraining and previous stages, leaving it vulnerable to distortion under new-task supervision and thereby causing forgetting.
We observe that the most pronounced drift tends to concentrate in vulnerable neighborhoods near the old-new semantic interface, where shared visual patterns are easily re-explained by new textual semantics.
To address this under an exemplar-free constraint, we propose Semantic Geometry Preservation for Continual Learning (SeGP-CL).
SeGP-CL first probes the drift-prone region by constructing a compact set of adversarial anchors with dual-targeted projected gradient descent (DPGD), which drives selected new-task seeds toward old-class semantics while remaining faithful in raw visual space.
During training, we preserve cross-modal structure by anchor-guided cross-modal geometry distillation (ACGD), and stabilize the textual reference frame across tasks via a lightweight text semantic-geometry regularization (TSGR).
In post-training stage, we estimate anchor-induced raw-space drift to transfer old visual prototypes and perform dual-path inference by fusing cross-modal and visual cues.
Extensive experiments on common continual learning benchmarks demonstrate that SeGP-CL consistently improves stability and forward transfer, achieving state-of-the-art performance while better preserving semantic geometry of VLMs. Code is available at: \url{https://github.com/chiyuan-IVIPLab/SeGP-CL}.
\end{abstract}


\section{Introduction}
\IEEEPARstart{C}{ontinual} learning (CL) aims to enable deep models to accumulate knowledge from a stream of tasks while mitigating catastrophic forgetting of previously acquired capabilities. With the emergence of large-scale pretrained vision-language models (VLMs) such as CLIP \cite{radford2021learning} and ALIGN \cite{jia2021scaling}, the CL paradigm is undergoing a notable shift: empowered by strong cross-modal representation alignment, VLMs can leverage stable and semantically explicit textual concepts to organize ever-growing downstream visual data, making them particularly suitable for class-incremental continual learning \cite{Chen2025CSTA,cheng2025achieving,ma2025joint,zhou2025learning} and related settings \cite{liang2025class,qiu2025mixtures,qiu2025null,liang2023new}.
\begin{figure}[!t]
  \centering
  \includegraphics[width=\linewidth]{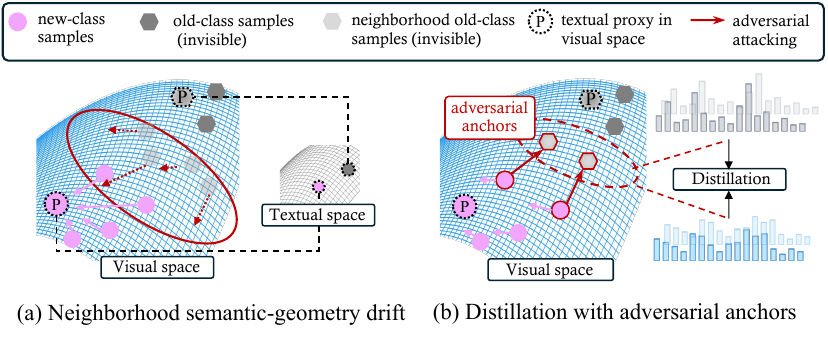}
  \vspace{-2.5em}
  \caption{Boundary vulnerability in VLM-based continual learning and our remedy.
  (a) Shared visual patterns near the old-new semantic interface are re-explained by new-task texts, leading to cross-modal semantic-geometry distortion and forgetting.
  (b) We construct adversarial anchors toward old-class semantics and perform anchor-guided cross-modal geometry distillation (ACGD) to constrain the drift in this vulnerable region.}
  \label{fig:intro-ab}
\end{figure}
Despite this promise, CL upon pretrained VLMs remains highly challenging. Early studies often adopt conservative strategies by attaching task-specific or class-specific components to an otherwise frozen VLM backbone \cite{zhang2024continual,zhou2025external,huang2025class,zhou2025learning}. These methods implicitly treat inter-task knowledge as primarily interfering, favoring isolation over transfer. However, such an assumption does not fully align with the general-purpose cross-modal structure learned by VLM pretraining. More recent efforts have shown progress by directly updating the VLM itself through parameter-efficient or structured adaptation schemes \cite{zhang2024overcoming,zhang2023slca,yu2024boosting}. Nevertheless, these approaches are largely general continual adaptation techniques and lack targeted modeling of cross-modal stability and knowledge transfer mechanisms. Another line of work begins to explicitly exploit the cross-modal priors of VLMs. In particular, these works exploit the rich representational capacity of the text modality (e.g., attributes, concepts, and semantic hierarchies) to facilitate anti-forgetting continual learning \cite{he2025desclip,yu2025language,hu2025hierarchical}. Yet, these methods still provide limited attention to \emph{preserving the established cross-modal geometry under the exemplar-free constraint}.

Recent studies have observed that abrupt representation variation induced by incremental updates can impair the cross-modal geometry of VLMs~\cite{suzuki2025dive,huang2025mind,zheng2023preventing}. However, existing solutions are often conservative, such as early stopping or regularization with reference data~\cite{changpinyo2021conceptual,deng2009imagenet}, and they struggle to precisely constrain the regions that are most susceptible to geometry distortion. Our key observation is that harmful drift does not occur uniformly in the embedding space. Instead, it tends to concentrate around the old-new semantic interface, where shared visual patterns are prone to being re-explained by newly introduced textual semantics. As illustrated in Fig.~\ref{fig:intro-ab}(a), once these shared patterns are pulled toward new semantics, the established visual-text alignment inherited from old tasks can be disrupted and lead to pronounced forgetting.

\begin{figure}[!t]
  \centering
  \includegraphics[width=0.5\textwidth]{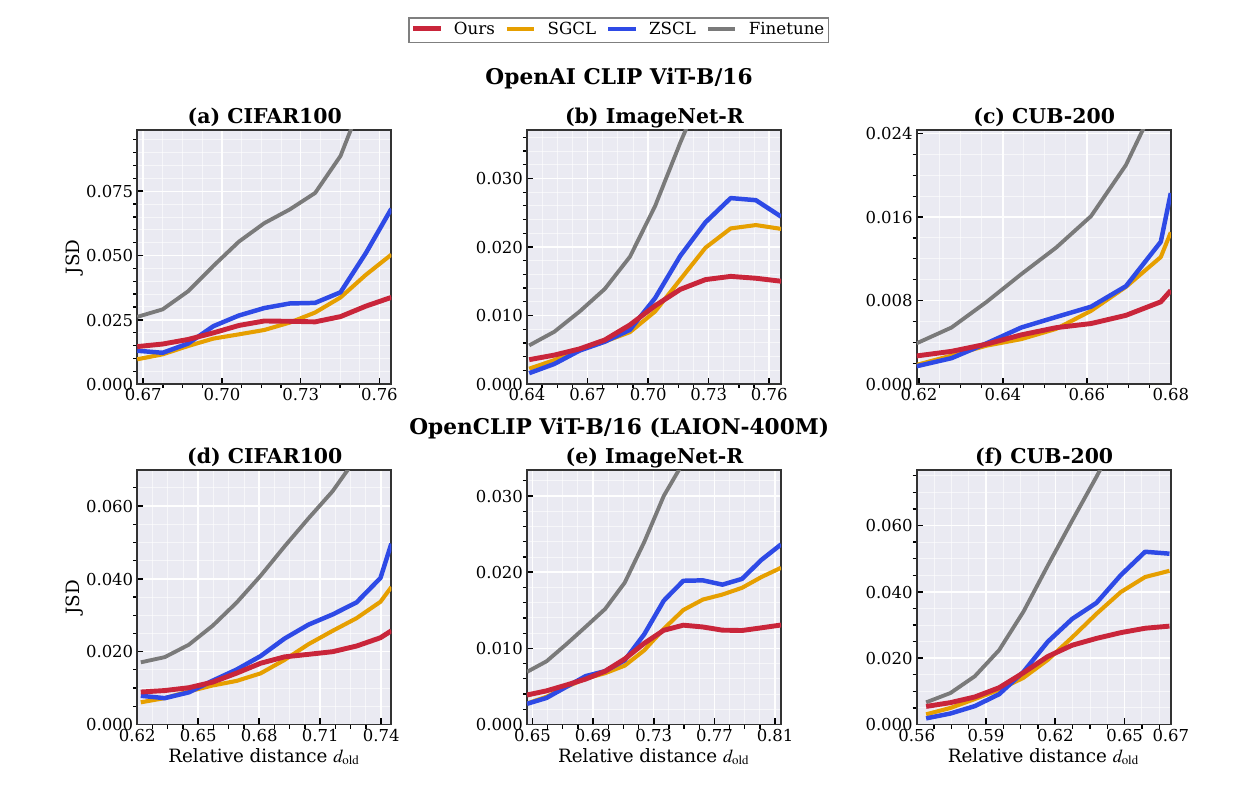}
  \vspace{-2em}
  \caption{Systematic evidence of boundary vulnerability. We report JSD-based cross-modal distributional drift with respect to the relative distance \(d_{\mathrm{old}}\) from the ground-truth old-class text embeddings, where \(d_{\mathrm{old}}\) is formally defined in Sec.~\ref{subsec:visualization}. Larger \(d_{\mathrm{old}}\) indicates that an old sample moves from the old-class semantic core toward the semantic boundary. Across different datasets, VLM backbones, and representative cross-modal-regularization CL methods, boundary regions consistently suffer larger drift, while our method reduces the drift in high-\(d_{\mathrm{old}}\) regions.}
  \label{fig:intro-exp}
\end{figure}

To examine this phenomenon more systematically, we measure the cross-modal distributional shift after incremental adaptation using the Jensen-Shannon divergence (JSD)~\cite{lin1991divergence} between the pre-update and post-update similarity distributions. As shown in Fig.~\ref{fig:intro-exp}, across multiple datasets, backbones, and representative cross-modal-regularization CL methods, e.g., ZSCL~\cite{zheng2023preventing} and SGCL~\cite{yu2024exploiting}, the JSD drift consistently becomes larger as old samples move farther away from their ground-truth old-class text embeddings, i.e., from old-class semantic cores toward semantic boundaries. This indicates that boundary-region drift is a general phenomenon rather than a single-case observation. Moreover, existing cross-modal regularization methods alleviate the drift to some extent but still exhibit amplified shifts in high-\(d_{\mathrm{old}}\) regions, while our method consistently reduces such boundary drift.

Inspired by findings on the adversarial robustness of VLMs~\cite{yu2024textguided,liu2025evaluating,chen2024attentionguided}, we leverage a practical insight: if tiny, targeted perturbations can quickly alter image-text relations, then such perturbations can be used constructively to expose and cover the most vulnerable neighborhoods of the old geometry. This motivates our design under the exemplar-free constraint: rather than synthesizing images that merely ``look like old classes''~\cite{wu2025synthetic,wang2025loraloop,qiu2024dual}, we propose to construct a small set of adversarial anchors that `push' selected new-task samples toward old-class semantic neighborhoods, thereby concentrating constraints on the boundary neighborhoods where updates are most prone to distort the established cross-modal geometry.

%To construct anchors, we propose Dual-targeted Projected Gradient Descent (DPGD), which pushes a new-task seed toward a selected old-class text embedding while constraining it to remain close to the corresponding visual prototype. The visual target compensates for the modality gap~\cite{liang2022mind} and prevents purely text-targeted attacks from generating unstable or visually irrelevant anchors. As a result, the anchors concentrate on vulnerable old-semantics neighborhoods close to the new-task semantics. We then apply Anchor-guided Cross-modal Geometry Distillation (ACGD) on these anchors to efficiently probe the drift-prone region and preserve the learned structure. 

%In addition, we introduce a Text Semantic Geometry Regularization (TSGR) to stabilize the textual semantic ``reference frame'' during continual updates. The reliability of cross-modal alignment depends not only on the visual-text relations at the trigger-region anchors, but also on the relative geometric structure among textual concepts (e.g., neighborhood relations and semantic hierarchies). If this structure is distorted across tasks, the semantic coordinates of old classes can be implicitly re-parameterized, causing old knowledge to degrade even when local distillation constraints are satisfied. TSGR constrains concept-wise relative relations through a key relational subgraph, preserving the core topology of the text semantic space with low overhead and providing a stable, transferable semantic reference for cross-modal geometric distillation.
To construct anchors, we propose Dual-targeted Projected Gradient Descent (DPGD), which pushes a new-task seed toward a selected old-class text embedding while constraining it to remain close to the corresponding visual prototype. The visual target compensates for the modality gap~\cite{liang2022mind} and prevents purely text-targeted attacks from generating unstable or visually irrelevant anchors. As a result, the anchors concentrate on vulnerable old-semantics neighborhoods close to the new-task semantics. We then apply Anchor-guided Cross-modal Geometry Distillation (ACGD) on these anchors to efficiently constrain the drift-prone region. Since the text encoder is also updated during continual adaptation, we further introduce Text Semantic Geometry Regularization (TSGR) to stabilize the textual semantic reference frame by preserving local relations among seen-class text embeddings, providing a reliable text-side basis for cross-modal geometry preservation.

Finally, we exploit the adversarial anchors to estimate and compensate the drift of old visual prototypes in the raw feature space under the exemplar-free constraint. Specifically, after finishing the current-task training, we measure the anchor-induced variations of raw visual features before and after adaptation, and use them to transfer the old-class prototypes so that the raw-space decision reference remains consistent with the updated visual encoder. The transferred prototypes are subsequently used as updated visual targets for the next-task DPGD procedure, and as complementary cues for dual-path prediction by fusing cross-modal logits with prototype-based visual logits. This is particularly important due to the modality gap \cite{liang2022mind, huang2025mind} in VLMs, where easy texts alone cannot fully characterize the visual space and discriminative raw visual patterns can provide additional evidence for robust inference. 
The contributions of this work can be summarized as follows:
\begin{itemize}
    \item We reveal that harmful cross-modal semantic-geometry drift in VLM continual learning is concentrated near the old-new semantic interface, and provide both empirical evidence and formal analysis for this boundary vulnerability.

    \item We propose SeGP-CL, an exemplar-free framework that constructs dual-targeted adversarial anchors to probe drift-prone boundary neighborhoods and preserves semantic geometry through ACGD and TSGR.

    \item We introduce anchor-induced prototype transfer to update old visual prototypes without storing old-task data, and perform dual-path inference by combining cross-modal semantics with raw visual cues.

    \item Extensive experiments on common continual learning benchmarks demonstrate state-of-the-art performance, with comprehensive analyses validating boundary-drift suppression and robust knowledge transfer.
\end{itemize}

\section{Related Work}
\label{sec:Related Work}

\subsection{Foundational Vision-language Models}
Large-scale pretrained vision-language models (VLMs) such as CLIP \cite{radford2021learning} and ALIGN \cite{jia2021scaling} are typically trained on web-scale image-text pairs with contrastive objectives, learning aligned representations between a visual encoder and a textual encoder. This large-scale pretraining yields strong zero-shot transfer, enabling open-vocabulary recognition via text prompts. For downstream tasks, VLMs can be adapted in a parameter-efficient manner via prompt learning \cite{zhou2022learning, zhou2022conditional}, adapter-based tuning \cite{li2024graphadapter,xin2024vmt}, and LoRA fine-tuning methods \cite{hu2022lora, zanella2024clip_lora}. However, these adaptation methods face severe forgetting challenges in multi-step continual learning scenarios.

\subsection{Continual Learning}
\label{sec:Continual Learning}
Continual learning studies incremental acquisition without catastrophic forgetting. Typical strategies include model expansion \cite{zhou2022model, dong2025CEAT}, knowledge distillation \cite{li2017learning}, and parameter regularization \cite{kirkpatrick2017overcoming,cheng2025achieving}. These solutions are generally developed for training-from-scratch small backbones, while recent work begins to incrementally inject knowledge into pretrained models. Representative directions include task-deconflicted prompt structures \cite{wang2022learning, smith2023coda}, adapter-based designs \cite{zhou2024expandable, yu2024boosting}, and LoRA-based continual learning \cite{liang2024inflora}. 

Due to the dual-tower cross-modal alignment structure of VLMs and the fact that textual concepts can provide semantically explicit class anchors, continual learning on VLMs has become increasingly popular in recent years. The key challenge lies in preserving the inherent cross-modal semantic geometry of VLMs as well as the alignment structure learned from previous tasks. A common direction is to introduce parameter-efficient task components, including learnable prompts \cite{wang2022learning, gao2024consistent} and adapters \cite{ zhou2025revisiting, yu2024boosting}. Beyond these adaptation methods, approaches exploiting textual priors (e.g., attribute/semantic structures) have shown effectiveness for exemplar-free class-incremental continual learning \cite{zhang2024continual, zhou2025external, hu2025hierarchical}. However, they typically require an external language expert (e.g., GPT-4) to obtain additional vision-related textual semantic information. Another line updates VLM parameters selectively to reduce generic knowledge loss \cite{zhang2024overcoming}.

Some works have highlighted that preserving the cross-modal geometry of VLMs is crucial for maintaining stable zero-shot transfer ability. Huang et al.~\cite{huang2025mind} show that aggressively narrowing the modality gap during continual updates may damage the inherent cross-modal structure, and propose modality-preservation and modality-compensation strategies. Suzuki et al.~\cite{suzuki2025dive} constrain cross-modal representation shifts on reference data to suppress structural collapse. Other works maintain VLM generality through reference-data-based distillation or monitoring~\cite{zheng2023preventing,yu2025select,zheng2024adaptwithoutforgetting}, but require additional reference data. In contrast, SGCL~\cite{yu2024exploiting} exploits a frozen text encoder to provide semantic soft labels and old-class cross-modal distillation on current-task images.

Different from these methods, we focus on the vulnerable old-new semantic interface rather than only global cross-modal preservation. We construct dual-targeted adversarial anchors to probe boundary neighborhoods, apply anchor-guided cross-modal geometry distillation to preserve old-class image-text similarity distributions in these regions, and further regularize the text semantic geometry because our textual encoder is also adapted. Thus, SeGP-CL preserves targeted semantic-interface geometry without old exemplars or external reference data.

\subsection{Adversarial Attacking on VLMs}

VLMs are vulnerable to small, structured perturbations that can rapidly alter image-text alignment, motivating adversarial attack and robustness studies~\cite{yu2024textguided,chen2024attentionguided,liu2025evaluating}. Existing PGD-style attacks usually maximize an adversarial objective to fool VLM prediction or disrupt cross-modal alignment, e.g., task-level co-attacks~\cite{zhang2022vlpattack}, set-level attacks~\cite{lu2023set}, and adversarial prompt tuning~\cite{zhang2024advpt}. Different from these attack-oriented methods, we use perturbations as geometry probes rather than attack samples. Specifically, DPGD moves selected new-task seeds toward old textual semantics while constraining them by old visual prototypes, thereby constructing compact anchors around the vulnerable old-new semantic interface for ACGD and prototype transfer.

\section{Methodology}
\label{sec:Methodology}

\begin{figure*}[htbp]
  \centering
  \includegraphics[width=\linewidth]{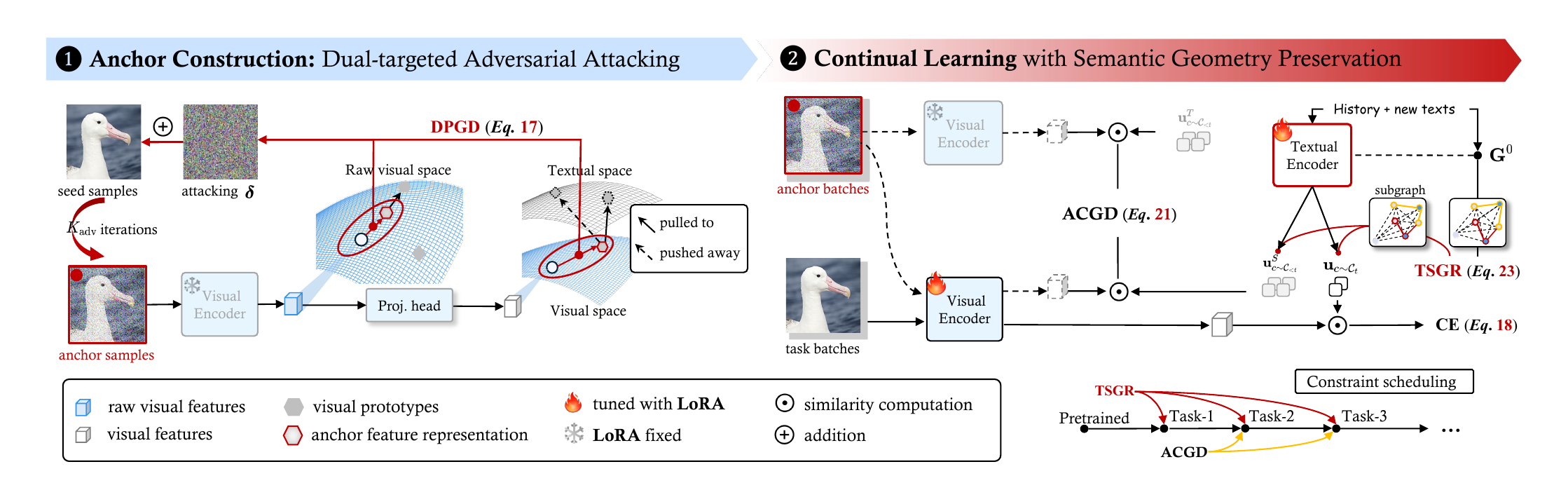}
  \vspace{-2.5em}
  \caption{Overview of proposed \textbf{SeGP-CL}. 1) Anchor construction: Dual-targeted projected gradient descent (DPGD) iteratively perturbs seed samples to synthesize adversarial anchors that are simultaneously guided in raw visual space and CLIP feature space. 2) Continual learning: A LoRA-tuned VLM is optimized on task batches with CE loss, while anchor batches and history texts impose semantic-geometry preservation via ACGD and TSGR.}
  \label{fig:pipeline}
\end{figure*}

\subsection{Preliminaries}
\label{subsec:prelim}
\textit{1) Continual Learning Formulation:}
A sequence of task datasets is denoted as \( \{\mathcal{D}_1, \mathcal{D}_2, \ldots, \mathcal{D}_T\} \).
During training on task \( \mathcal{D}_t \), access to previous data \( \{\mathcal{D}_1,\ldots,\mathcal{D}_{t-1}\} \) is restricted.
In class-incremental continual learning, task \(t\) introduces a disjoint class set \(\mathcal{C}_t\), and we define
\(
\mathcal{C}_{<t}=\bigcup_{i=1}^{t-1}\mathcal{C}_i
\)
and
\(
\mathcal{C}_{\le t}=\mathcal{C}_{<t}\cup\mathcal{C}_t
\).
The training set for task $t$ only includes:
\begin{equation}
\mathcal{D}_t=\{(\mathbf{x},y)\mid y\in\mathcal{C}_t\}.
\end{equation}

\textit{2) CLIP Representations:}
As a representative foundational vision-language model (VLM), CLIP \cite{radford2021learning} consists of a visual encoder
\(\mathbf{F}(\cdot)\) and a textual encoder \(\mathbf{G}(\cdot)\) that map images and texts into an aligned embedding space.
The visual encoder is decoupled into a raw extractor and a projection head $\mathbf{P}_{r\rightarrow v}\!\left(\cdot\right)$:
\begin{equation}
\mathbf{r}(\mathbf{x})=\mathbf{F}^{*}(\mathbf{x})\in\mathbb{R}^{D},
\qquad
\mathbf{v}(\mathbf{x})=\mathbf{P}_{r\rightarrow v}\!\left(\mathbf{r}(\mathbf{x})\right)\in\mathbb{R}^{d}.
\end{equation}
We use \(\ell_2\)-normalized features
\begin{equation}
\bar{\mathbf{r}}(\mathbf{x})=
\frac{\mathbf{r}(\mathbf{x})}{\|\mathbf{r}(\mathbf{x})\|_2},
\qquad
\bar{\mathbf{v}}(\mathbf{x})=
\frac{\mathbf{v}(\mathbf{x})}{\|\mathbf{v}(\mathbf{x})\|_2}.
\end{equation}
For class \(c\), the text embedding is constructed by prompting ($\mathbf{pr.}$) the textual encoder using the template such as ``A photo of a [\textit{CLASS}].":
\begin{equation}
\mathbf{u}_{c}=
\frac{\mathbf{G}(\mathbf{pr.}(c))}{\|\mathbf{G}(\mathbf{pr.}(c))\|_2}
\in\mathbb{R}^{d}.
\label{eq:text_embedding}
\end{equation}
CLIP predicts class \(c\) by the cross-modal logits $s^{\mathrm{clip}}(\mathbf{x},c) = \bar{\mathbf{v}}(\mathbf{x})^\top \mathbf{u}_{c},$ and for a class set \(\mathcal{S}\subseteq\mathcal{C}_{\le t}\), the probability is
\begin{equation}
p^{\mathrm{clip}}(y=c\mid \mathbf{x};\mathcal{S})
=
\frac{\exp\!\left(s^{\mathrm{clip}}(\mathbf{x},c)/\tau\right)}
{\sum_{j\in\mathcal{S}}\exp\!\left(s^{\mathrm{clip}}(\mathbf{x},j)/\tau\right)},
\end{equation}
where $\tau$ is a temperature that controls the distribution softness.
\textit{3) Visual Branch:}
To complement cross-modal semantics with discriminative raw visual patterns, we maintain one normalized
raw visual space prototype per class \cite{zhou2025external,li2025bofa}, \(\boldsymbol{\mu}_{c}\in\mathbb{R}^{D}\) with \(\|\boldsymbol{\mu}_{c}\|_2=1\).
Given \(\bar{\mathbf{r}}(\mathbf{x})\), the prototype-branch logits is $s^{v}(\mathbf{x},c) = \bar{\mathbf{r}}(\mathbf{x})^\top \boldsymbol{\mu}_{c},$
and the corresponding probability over \(\mathcal{S}\) is
\begin{equation}
p^{v}(y=c\mid \mathbf{x};\mathcal{S})
=
\frac{\exp\!\left(s^{v}(\mathbf{x},c)/\tau\right)}
{\sum_{j\in\mathcal{S}}\exp\!\left(s^{v}(\mathbf{x},j)/\tau\right)}.
\label{eq:V. pred}
\end{equation}

\subsection{Formal Analysis of Boundary Vulnerability}
\label{subsec:boundary_analysis}

We further analyze why downstream adaptation tends to distort the old-new semantic interface. At task \(t\), the supervised loss is computed only over the current class set \(\mathcal{C}_t\). For \((\mathbf{x},y)\in\mathcal{D}_t\), the feature-level descent direction of the cross-modal CE loss is
\begin{equation}
\begin{aligned}
-\nabla_{\bar{\mathbf{v}}(\mathbf{x})}\ell_t(\mathbf{x},y)
=
\frac{1}{\tau}
\left(
\mathbf{u}_{y}
-
\sum_{c\in\mathcal{C}_t}
p^{\mathrm{clip}}(y=c\mid \mathbf{x};\mathcal{C}_t)
\mathbf{u}_{c}
\right).
\end{aligned}
\label{eq:local_ce_direction}
\end{equation}
This indicates that current-task adaptation is driven by current-class textual embeddings, while old-class visual-text relations are not explicitly preserved by the supervised objective.

To analyze the old-new relation, we make the dependence of the CLIP similarity on the trainable parameters \(\theta\) explicit. For an old class \(o\in\mathcal{C}_{<t}\), a new class \(y\in\mathcal{C}_t\), and a probe sample \(\mathbf{z}\), define \(g_t=\nabla_{\theta}\ell_t(\mathbf{x},y)\). The old-new margin and its first-order change after one update are
\begin{equation}
\begin{aligned}
m_{o,y}(\mathbf{z};\theta)
&=
s^{\mathrm{clip}}(\mathbf{z},o;\theta)
-
s^{\mathrm{clip}}(\mathbf{z},y;\theta),\\
\Delta m_{o,y}(\mathbf{z})
&\approx
-\eta
\left\langle
\nabla_{\theta}m_{o,y}(\mathbf{z};\theta),
g_t
\right\rangle .
\end{aligned}
\label{eq:margin_coupling}
\end{equation}
The inner product measures the coupling between the old-new relation of \(\mathbf{z}\) and the current-task gradient. Interface samples usually share visual attributes with new classes and have non-negligible affinity to new textual embeddings, making their margins more susceptible to current-task updates.

The interface is also more sensitive to similar margin perturbations. For the pair \((o,y)\), let \(P_{o,y}(\mathbf{z})\) be the pairwise probability of assigning \(\mathbf{z}\) to the old class \(o\):
\begin{equation}
\begin{aligned}
P_{o,y}(\mathbf{z})
&=
\sigma\!\left(
m_{o,y}(\mathbf{z};\theta)/\tau
\right),\\
\Delta P_{o,y}(\mathbf{z})
&\approx
\frac{1}{\tau}
P_{o,y}(\mathbf{z})
\left(1-P_{o,y}(\mathbf{z})\right)
\Delta m_{o,y}(\mathbf{z}).
\end{aligned}
\label{eq:prob_sensitivity}
\end{equation}
The factor \(P_{o,y}(1-P_{o,y})\) is maximized when \(m_{o,y}(\mathbf{z};\theta)\approx0\), i.e., near the old-new semantic interface, and becomes small in saturated core regions. Therefore, boundary neighborhoods are vulnerable because they combine stronger gradient coupling and higher distributional sensitivity. This explains the amplified JSD around the semantic interface observed in Fig.~\ref{fig:intro-exp}, and motivates DPGD to construct anchors that probe these drift-prone regions for preservation.

\subsection{Overview}
\label{subsec:overview}
\begin{figure}
    \centering
    \includegraphics[width=\linewidth]{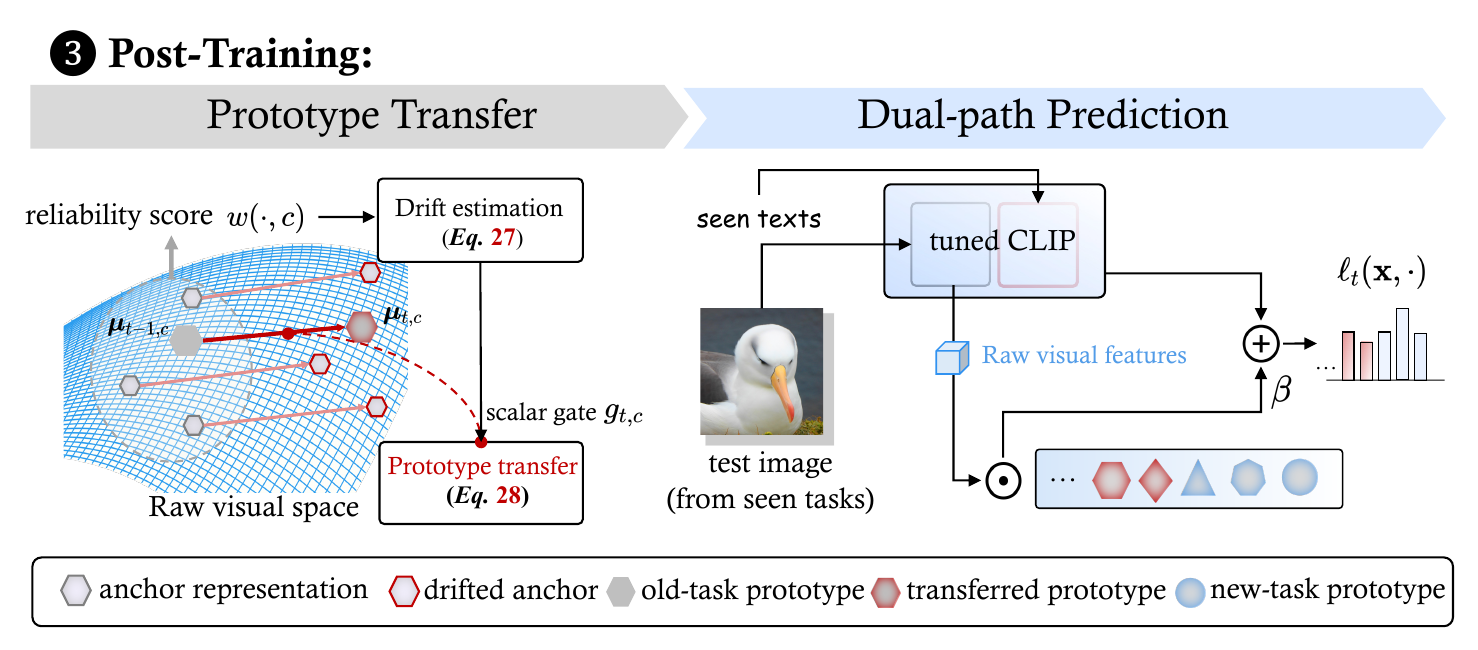}
    \vspace{-2em}
    \caption{Post-Training: We estimate raw-space drift and transfer visual prototypes for old classes, then perform dual-path inference by ensembling logits from visual and CLIP branches.}
    \label{fig:after-train}
\end{figure}

As illustrated in Fig.~\ref{fig:pipeline} and Fig.~\ref{fig:after-train}, our exemplar-free continual learning framework, \textbf{Se}mantic \textbf{G}eometry \textbf{P}reservation (SeGP-CL), follows a three-stage procedure for each incremental task. In \textit{Stage-I: Anchor Construction}, before learning task \(t\), we freeze the model from the previous task as a teacher snapshot, i.e., \((\mathbf{F}^{T},\mathbf{G}^{T})=(\mathbf{F}_{t-1},\mathbf{G}_{t-1})\), select old-related seeds from the current task data \(\mathcal{D}_t\), and construct a compact anchor set \(\mathcal{A}_t\) via Dual-targeted Projected Gradient Descent (DPGD). These anchors are pushed toward old-class textual semantics while remaining close to the corresponding old visual prototypes, thereby probing the vulnerable old-new semantic interface. In \textit{Stage-II: Continual Learning with SeGP}, we optimize the student with new-class supervision and preserve prior cross-modal structure by applying Anchor-guided Cross-modal Geometry Distillation (ACGD) on \(\mathcal{A}_t\) over \(\mathcal{C}_{<t}\), together with a lightweight Text Semantic Geometry Regularization (TSGR) to stabilize the textual semantic reference frame. In \textit{Stage-III: Post-Training}, after finishing task-\(t\) training, we estimate anchor-induced drift in the raw visual space to transfer old prototypes, and perform dual-path prediction by fusing logits from the CLIP branch and the prototype-based visual branch for robust inference over all seen classes. To complement the overview above, Algorithm~\ref{alg:segp} summarizes the three-stage training and inference procedure of SeGP-CL in each incremental task.

\begin{algorithm}[t]
\caption{\textsc{SeGP-CL} Procedure.}
\label{alg:segp}
\footnotesize
\KwInput{pretrained CLIP \((\mathbf{F}_0,\mathbf{G}_0)\); task stream \(\{(\mathcal{D}_t,\mathcal{C}_t)\}_{t=1}^{T}\); prompts \(\mathbf{pr.}(\cdot)\).}
\KwOutput{adapted model \((\mathbf{F}_{T},\mathbf{G}_{T})\); prototype memory \(\mathcal{M}_{T}\).}
\KwInit{insert LoRA into \((\mathbf{F}_0,\mathbf{G}_0)\); \(\mathcal{C}_{\le0}\leftarrow\emptyset\), \(\mathcal{M}_{0}\leftarrow\emptyset\).}

\For{task \(t=1\) to \(T\)}{
    Update seen classes: \(\mathcal{C}_{<t}\leftarrow\mathcal{C}_{\le t-1}\), \(\mathcal{C}_{\le t}\leftarrow\mathcal{C}_{<t}\cup\mathcal{C}_{t}\)\;
    Freeze teacher: \((\mathbf{F}^{T},\mathbf{G}^{T})\leftarrow(\mathbf{F}_{t-1},\mathbf{G}_{t-1})\), \(\mathcal{A}_{t}\leftarrow\emptyset\)\;

    \textcolor{stagered}{\(\blacktriangleright\) \textit{Stage-I: Anchor Construction}}\;
    \ForEach{\(c\in\mathcal{C}_{<t}\)}{
        Compute textual embeddings: \(\mathbf{u}^{T}_{c}\) by Eq.~\eqref{eq:text_embedding}\;
        Select seeds: \(\mathcal{I}_{c}\leftarrow\mathrm{TopK}(Q(\mathbf{x},c))\) by Eq.~\eqref{eq:seed_score}\;
        Generate anchors: \(\mathcal{A}_{t}^{c}\leftarrow\mathrm{DPGD}(\mathcal{I}_{c},\mathbf{u}^{T}_{c},\boldsymbol{\mu}_{t-1,c})\) by Eqs.~\eqref{eq:adv_obj}--\eqref{eq:pgd}\;
        Collect anchors: \(\mathcal{A}_{t}\leftarrow\mathcal{A}_{t}\cup\mathcal{A}_{t}^{c}\)\;
    }

    \textcolor{stagered}{\(\blacktriangleright\) \textit{Stage-II: Continual Learning with SeGP}}\;
    Construct text subgraphs: \(\{\mathcal{N}_{k}(c)\}_{c\in\mathcal{C}_{t}}\) from \(\mathbf{G}^{0}\)\;
    \ForEach{mini-batch}{
        Compute CE loss: \(\mathcal{L}_{\mathrm{cls}}\leftarrow\mathrm{CE}(\mathcal{D}_{t},\mathcal{C}_{t})\) by Eq.~\eqref{eq:cls_loss}\;
        Compute ACGD loss:\;
        \(\mathcal{L}_{\mathrm{ACGD}}\leftarrow
        \mathrm{CGD}(\mathcal{A}_{t},\mathcal{C}_{<t};
        \mathbf{F}^{T},\mathbf{G}^{T},\mathbf{F}^{S},\mathbf{G}^{S})\) by Eq.~\eqref{eq:acgd_loss}\;
        Compute TSGR loss:\;
        \(\mathcal{L}_{\mathrm{GR}}\leftarrow
        \mathrm{TSGR}(\{\mathcal{N}_{k}(c)\}_{c\in\mathcal{C}_{t}};
        \mathbf{G}^{0},\mathbf{G}^{S})\) by Eq.~\eqref{eq:gr_loss}\;
        Update student: \((\mathbf{F}^{S},\mathbf{G}^{S})\) by Eq.~\eqref{eq:overall_loss}\;
    }

    Obtain adapted model: \((\mathbf{F}_{t},\mathbf{G}_{t})\leftarrow(\mathbf{F}^{S},\mathbf{G}^{S})\)\;

    \textcolor{stagered}{\(\blacktriangleright\) \textit{Stage-III: Post-Training}}\;
    Estimate new prototypes: \(\boldsymbol{\mu}_{t,c}\), \(c\in\mathcal{C}_{t}\), by Eq.~\eqref{eq:new_proto}\;
    Transfer old prototypes: \(\boldsymbol{\mu}_{t,c}\), \(c\in\mathcal{C}_{<t}\), by Eq.~\eqref{eq:old_proto_transfer}\;
    Store memory: \(\mathcal{M}_{t}\leftarrow\{\boldsymbol{\mu}_{t,c}\}_{c\in\mathcal{C}_{\le t}}\)\;
    Inference: predict \(\hat{y}\) over \(\mathcal{C}_{\le t}\) by Eqs.~\eqref{eq:dual_path_logits}--\eqref{eq:dual_path_pred}\;
}
\end{algorithm}

\subsection{Anchor Construction}
\label{subsec:anchor}

To concentrate supervision on the most vulnerable old-new semantic interface under the exemplar-free setting, we construct a compact set of adversarial anchors that probe the boundary neighborhoods of the old cross-modal geometry. The key idea is to (i) fix a stable semantic reference that faithfully reflects the pre-update old decision geometry, and (ii) start from incoming samples that already exhibit strong affinity to old semantics, so that a small perturbation is sufficient to reach the drift-prone regions. In addition, we adopt iterative projected gradient descent (PGD) \cite{madry2017towards} as the anchor generator because it provides an efficient exploration strategy for drift-prone interface neighborhoods under a bounded perturbation budget. Specifically, let $L(\mathbf{x},c)$ denote a teacher-defined objective whose decrease drives $\mathbf{x}$ toward the old semantic region of class $c$ (in our case, $L=\mathcal{L}_{\mathrm{adv}}$ in Eq.~\eqref{eq:adv_obj}). For a single perturbation step $\Delta$ with $\|\Delta\|_{\infty}\le\gamma$, a first-order approximation gives
\begin{equation}
L(\mathbf{x}+\Delta,c)\approx L(\mathbf{x},c)+\left\langle\nabla_{\mathbf{x}}L(\mathbf{x},c),\Delta\right\rangle .
\label{eq:taylor}
\end{equation}
By H\"older's inequality, for any feasible $\Delta$,
\begin{equation}
\left\langle\nabla_{\mathbf{x}}L,\Delta\right\rangle
\ge
-\|\nabla_{\mathbf{x}}L\|_{1}\cdot\|\Delta\|_{\infty},
\label{eq:holder}
\end{equation}
and the bound is achieved by the sign step $\Delta^\star=-\gamma\,\mathrm{sign}(\nabla_{\mathbf{x}}L)$. Therefore, the sign-gradient update yields the maximum instantaneous decrease of the linearized objective among all $\ell_\infty$-bounded steps, i.e., it is the steepest-descent direction under the $\ell_\infty$ constraint. Iterating such locally optimal steps efficiently drives seed samples toward and across the teacher-induced old semantic interface, thereby probing the most vulnerable boundary neighborhoods with a small number of iterations.

Concretely, before learning task $t$, we freeze the model from the previous task as a teacher snapshot $(\mathbf{F}^{T},\mathbf{G}^{T})=(\mathbf{F}_{t-1},\mathbf{G}_{t-1})$. This teacher provides pre-update cross-modal semantics, ensuring that anchor construction is guided by the old decision geometry rather than being contaminated by the ongoing update. Given the new-task dataset $\mathcal{D}_t$, we select a small set of seeds for each old class $c\in\mathcal{C}_{<t}$ by measuring how strongly a new sample is already aligned with the old text embedding under the teacher. Specifically, for $\mathbf{x}\in\mathcal{D}_t$, we compute the teacher cross-modal similarity
\begin{equation}
\mathrm{Q}(\mathbf{x},c)=\bar{\mathbf{v}}^{T}(\mathbf{x})^\top \mathbf{u}^{T}_{c},
\label{eq:seed_score}
\end{equation}
where $\bar{\mathbf{v}}^{T}(\mathbf{x})=
\frac{\mathbf{v}^{T}(\mathbf{x})}{\left\|\mathbf{v}^{T}(\mathbf{x})\right\|_2},\mathbf{u}^{T}_{c}=
\frac{\mathbf{G}^{T}(\mathbf{pr.}(c))}{\left\|\mathbf{G}^{T}(\mathbf{pr.}(c))\right\|_2}.$
We retain the $K_{\mathrm{seed}}$ top-ranked samples as $\mathcal{I}_c$, which serve as initialization points for adversarial perturbation. Intuitively, these seeds capture shared visual patterns that already lie close to old semantics in the teacher embedding space, making them ideal starting points to probe the boundary neighborhoods that are most vulnerable to drift. Starting from a seed $\mathbf{x}\in\mathcal{I}_c$, we generate an adversarial anchor $\mathbf{x}^{\mathrm{adv}}=\mathbf{x}+\boldsymbol{\delta}$ within an $\ell_\infty$ budget by optimizing a text-targeted objective:
\begin{equation}
\min_{\|\boldsymbol{\delta}\|_\infty\le \epsilon}\;
\mathcal{L}_{\mathrm{adv}}(\mathbf{x}+\boldsymbol{\delta},c),
\label{eq:adv_obj_txt}
\end{equation}
where $\mathcal{L}_{\mathrm{adv}}(\mathbf{x}+\boldsymbol{\delta},c)$ enforces a relative preference for class $c$ among all old classes, explicitly pushing the perturbed sample toward the teacher's old semantic region:
\begin{equation}
\mathcal{L}_{\mathrm{adv}}(\mathbf{x}+\boldsymbol{\delta},c)
=
-\log
\frac{
\exp\!\left(\bar{\mathbf{v}}^{T}(\mathbf{x}+\boldsymbol{\delta})^\top \mathbf{u}^{T}_{c}/\tau\right)
}{
\sum_{j\in\mathcal{C}_{<t}}
\exp\!\left(\bar{\mathbf{v}}^{T}(\mathbf{x}+\boldsymbol{\delta})^\top \mathbf{u}^{T}_{j}/\tau\right)
}.
\label{eq:txt_adv}
\end{equation}
However, purely text-targeted perturbations can become unreliable in the presence of the modality gap \cite{liang2022mind,huang2025mind}, since the geometric structures of the visual and textual embedding spaces are not perfectly corresponding. As a result, driving an image embedding toward a target text embedding alone may match the direction of textual semantics while deviating from class-consistent visual patterns, yielding unstable or visually implausible anchors. We therefore introduce a visual anchoring term in the raw feature space, which pulls the anchor toward the old raw-space visual prototype of class $c$:
\begin{equation}
\mathcal{L}_{\mathrm{v\text{-}adv}}(\mathbf{x}+\boldsymbol{\delta},c)
=
1-\bar{\mathbf{r}}^{T}(\mathbf{x}+\boldsymbol{\delta})^\top \boldsymbol{\mu}_{t-1,c},
\label{eq:v_adv}
\end{equation}
where $\bar{\mathbf{r}}^{T}(\mathbf{x})=\frac{\mathbf{r}^{T}(\mathbf{x})}{\left\|\mathbf{r}^{T}(\mathbf{x})\right\|_2}.$ Combining the original objective in Eq.~\eqref{eq:txt_adv} with the visual anchoring term in Eq.~\eqref{eq:v_adv}, we obtain the corrected dual-target objective:
\begin{equation}
\min_{\|\boldsymbol{\delta}\|_\infty\le \epsilon}\;
\mathcal{L}_{\mathrm{adv}}^{\prime}(\mathbf{x}+\boldsymbol{\delta},c)
=
\mathcal{L}_{\mathrm{adv}}(\mathbf{x}+\boldsymbol{\delta},c)
+
\lambda_p\,\mathcal{L}_{\mathrm{v\text{-}adv}}(\mathbf{x}+\boldsymbol{\delta},c),
\label{eq:adv_obj}
\end{equation}
where $\lambda_p$ is the correction coefficient. We solve Eq.~\eqref{eq:adv_obj} with standard PGD using step size $\gamma$:
\begin{equation}
\boldsymbol{\delta}^{(k+1)}
=
\Pi_{\|\boldsymbol{\delta}\|_\infty\le \epsilon}
\Big(
\boldsymbol{\delta}^{(k)}
-\gamma\,\mathrm{sign}\big(
\nabla_{\boldsymbol{\delta}}\mathcal{L}^{\prime}_{\mathrm{adv}}(\mathbf{x}+\boldsymbol{\delta}^{(k)},c)
\big)
\Big),
\label{eq:pgd}
\end{equation}
where $\Pi_{\|\boldsymbol{\delta}\|_\infty\le \epsilon}(\cdot)$ projects onto the $\ell_\infty$ ball of radius $\epsilon$. Repeating the update for $K_{\mathrm{adv}}$ iterations yields anchors that are confidently mapped to old semantics in the teacher embedding space while remaining visually faithful to the incoming distribution. We collect all generated anchors into $\mathcal{A}_t$, and record each anchor's associated old class for drift estimation and subsequent tasks.

\subsection{Continual Learning with Semantic-Geometry Preservation}
\label{subsec:train}

Given the anchor set \(\mathcal{A}_t\) constructed in Section~\ref{subsec:anchor}, we update the student model for task \(t\) by jointly balancing new-task acquisition and semantic-geometry preservation. We first optimize standard cross-entropy on the incoming label space \(\mathcal{C}_t\) to fit the new supervision:
\begin{equation}
\mathcal{L}_{\mathrm{cls}}
=
\mathbb{E}_{(\mathbf{x},y)\sim\mathcal{D}_t}
\left[-\log p^{\mathrm{clip}}_{t}(y\mid \mathbf{x};\mathcal{C}_t)\right].
\label{eq:cls_loss}
\end{equation}
However, optimizing \(\mathcal{L}_{\mathrm{cls}}\) alone can pull shared visual patterns toward new semantics and induce destructive drift around the old-new interface. We therefore preserve the established cross-modal geometry by distilling the teacher’s old-class distribution on anchors \(\mathbf{x}^{adv}\in\mathcal{A}_t\).
Here \(s^{T}(\mathbf{x},c)\) and \(s^{S}(\mathbf{x},c)\) denote the cross-modal CLIP logits evaluated by the frozen teacher snapshot \((\mathbf{F}^{T},\mathbf{G}^{T})\) and the current student \((\mathbf{F}^{S},\mathbf{G}^{S})\), respectively, following the same form as \(s^{\mathrm{clip}}(\mathbf{x},c)\) in Section~\ref{subsec:prelim}. Concretely,
\begin{equation}
\begin{aligned}
\text{teacher: }
s^{T}(\mathbf{x},c)
&=
\Bigg(
\frac{\mathbf{v}^{T}(\mathbf{x})}{\|\mathbf{v}^{T}(\mathbf{x})\|_2}
\Bigg)^\top
\underbrace{
\Bigg(
\frac{\mathbf{G}^{T}(\mathbf{pr.}(c))}{\|\mathbf{G}^{T}(\mathbf{pr.}(c))\|_2}
\Bigg)
}_{\text{snapshot textual embedding }\mathbf{u}^{T}_{c}},\\
\text{student: }
s^{S}(\mathbf{x},c)
&=
\Bigg(
\frac{\mathbf{v}^{S}(\mathbf{x})}{\|\mathbf{v}^{S}(\mathbf{x})\|_2}
\Bigg)^\top
\underbrace{
\Bigg(
\frac{\mathbf{G}^{S}(\mathbf{pr.}(c))}{\|\mathbf{G}^{S}(\mathbf{pr.}(c))\|_2}
\Bigg)
}_{\text{textual embedding }\mathbf{u}^{S}_{c}} ,
\end{aligned}
\end{equation}
for \(c\in \mathcal{C}_{<t}\). We then define the temperature-scaled teacher/student distributions using the anchors $\mathbf{x}^{adv}$ on \(\mathcal{C}_{<t}\) as
\begin{equation}
\begin{aligned}
\text{teacher:}
\pi_T^{\tau_A}(c\mid\mathbf{x}^{adv})
&=
\frac{\exp\!\big(s^{T}(\mathbf{x}^{adv},c)/\tau_A\big)}
{\sum_{j\in\mathcal{C}_{<t}}\exp\!\big(s^{T}(\mathbf{x}^{adv},j)/\tau_A\big)},\\
\text{student:}
\pi_S^{\tau_A}(c\mid\mathbf{x}^{adv})
&=
\frac{\exp\!\big(s^{S}(\mathbf{x}^{adv},c)/\tau_A\big)}
{\sum_{j\in\mathcal{C}_{<t}}\exp\!\big(s^{S}(\mathbf{x}^{adv},j)/\tau_A\big)},
\end{aligned}
\end{equation}
and minimize the Kullback-Leibler (KL) divergence \cite{kullback1951information} to obtain anchor-guided cross-modal geometry distillation (ACGD):
\begin{equation}
\mathcal{L}_{\mathrm{ACGD}}
=
{\tau_A}^2\,
\mathbb{E}_{\mathbf{x}^{adv}\sim\mathcal{A}_t}
\left[
\mathrm{KL}\big(\pi_T^{\tau_A}(\cdot\mid\mathbf{x}^{adv}) \,\|\, \pi_S^{\tau_A}(\cdot\mid\mathbf{x}^{adv})\big)
\right],
\label{eq:acgd_loss}
\end{equation}
where $\tau_A$ is a distillation temperature to control the distribution softness.

Beyond local cross-modal constraints, stable continual learning also requires a consistent textual semantic reference frame across tasks. Otherwise, the relative geometry among text concepts may drift and implicitly re-parameterize old semantics even if cross-modal distillation is satisfied. We therefore introduce Text Semantic-Geometry Regularization (TSGR). Concretely, at the beginning of task $t$, we obtain a fixed reference text encoder $\mathbf{G}^{0}$ by resetting the LoRA parameters in the current textual encoder, which recovers a stable textual embedding space shared across tasks. Based on this reference frame, we define a reference subgraph distribution in the reference text space and match it to the counterpart induced by the current student $\mathbf{G}^{S}$. Specifically, we construct the neighbor subgraph under the reference space induced by $\mathbf{G}^{0}$: for each new class $c\in\mathcal{C}_t$, we first compute the reference text embeddings for all seen classes $j\in\mathcal{C}_{\le t}$ as $\mathbf{u}^0_j=\frac{\mathbf{G}^{0}(\mathbf{pr.}(j))}{\|\mathbf{G}^{0}(\mathbf{pr.}(j))\|_2}$, and define $\mathcal{N}_k(c)$ as the indices of the $k$ nearest neighbors of $\mathbf{u}^0_c$ in this reference space (excluding $c$), i.e., the top-$k$ classes with the largest cosine similarity $(\mathbf{u}^0_{c})^\top \mathbf{u}^0_{j}$. Importantly, $\mathcal{N}_k(c)$ is computed once at the beginning of task $t$ using $\mathbf{G}^{0}$ and kept fixed during training, avoiding contamination from the evolving student text space. We then compute the current text embeddings as $\mathbf{u}^{S}_{j}=\frac{\mathbf{G}^{S}(\mathbf{pr.}(j))}{\|\mathbf{G}^{S}(\mathbf{pr.}(j))\|_2}$, and define the reference and student subgraph distributions on the same neighbor set as

\begin{equation}
\begin{alignedat}{2}
\text{reference:}
&\varphi_0(j\mid c)
&&=
\frac{
\exp\!\left((\mathbf{u}^0_{c})^\top \mathbf{u}^0_{j}/\tau_T\right)
}{
\sum_{\ell\in\mathcal{N}_k(c)}
\exp\!\left((\mathbf{u}^0_{c})^\top \mathbf{u}^0_{\ell}/\tau_T\right)
},
\\
\text{student: }
&\varphi_S(j\mid c)
&&=
\frac{
\exp\!\left((\mathbf{u}^{S}_{c})^\top \mathbf{u}^{S}_{j}/\tau_T\right)
}{
\sum_{\ell\in\mathcal{N}_k(c)}
\exp\!\left((\mathbf{u}^{S}_{c})^\top \mathbf{u}^{S}_{\ell}/\tau_T\right)
}.
\end{alignedat}
\end{equation}

where $ j\in\mathcal{N}_k(c)$, and $\tau_T$ is the temperature parameter. TSGR minimizes the KL divergence over all selected subgraphs:
\begin{equation}
\mathcal{L}_{\mathrm{GR}}
=
\frac{1}{|\mathcal{C}_t|}
\sum_{c\in\mathcal{C}_t}
\mathrm{KL}\big(\varphi_0(\cdot\mid c)\,\|\,\varphi_S(\cdot\mid c)\big).
\label{eq:gr_loss}
\end{equation}
Here, we apply TSGR only to the neighborhood subgraphs rooted at the new classes $c\in\mathcal{C}_t$, rather than enforcing a global relational constraint over all seen concepts. This choice is motivated by both effectiveness and efficiency. First, text embeddings in VLM-based continual learning are typically more stable than visual ones \cite{kang2025dynamic}, and the optimization at task $t$ primarily perturbs the newly introduced class texts and their nearby semantic neighborhood; thus, geometric distortion is more likely to arise locally around these new nodes. Second, TSGR is implemented via local $k$-NN subgraph matchings, whose cost scales with the number of rooted subgraphs. Restricting the constraint to $|\mathcal{C}_t|$ new-class roots yields $\mathcal{O}(|\mathcal{C}_t|k)$ similarity terms per step, which is substantially cheaper than the global $\mathcal{O}(|\mathcal{C}_{\le t}|k)$ overhead in class-incremental learning where $|\mathcal{C}_t|\ll|\mathcal{C}_{\le t}|$.

Combining the above components, the overall objective for task $t$ is given by
\begin{equation}
\min_{\boldsymbol{\theta}^{S}_{\mathrm{LoRA}}}\ \mathcal{L}^t_{\mathrm{CL}}
=
\mathcal{L}_{\mathrm{cls}}
+
\lambda_{\mathrm{ACGD}}\mathcal{L}_{\mathrm{ACGD}}
+
\lambda_{\mathrm{GR}}\mathcal{L}_{\mathrm{GR}},
\label{eq:overall_loss}
\end{equation}
where $\boldsymbol{\theta}^{S}_{\mathrm{LoRA}}$ denotes the LoRA parameters inserted into the student model $(\mathbf{F}^{S},\mathbf{G}^{S})$ at task $t$.

\subsection{Post-Training Stage}
\label{subsec:after}

\textit{1) Anchor-induced Prototype Transfer:}
After task-\(t\) training, we obtain the updated model \((\mathbf{F}_{t},\mathbf{G}_{t})\) and update visual prototypes to keep the raw-space decision references compatible with the adapted visual encoder.
For new classes \(c\in\mathcal{C}_t\), the prototype is directly estimated from the available task-\(t\) data:
\begin{equation}
\boldsymbol{\mu}_{t,c}
=
\frac{
\sum_{(\mathbf{x},y)\in\mathcal{D}_t,\;y=c}
\bar{\mathbf{r}}_{t}(\mathbf{x})
}{
\Big\|
\sum_{(\mathbf{x},y)\in\mathcal{D}_t,\;y=c}
\bar{\mathbf{r}}_{t}(\mathbf{x})
\Big\|_2
},
\quad \forall c\in\mathcal{C}_t,
\label{eq:new_proto}
\end{equation}
where \(\bar{\mathbf{r}}_{t}(\mathbf{x})=\mathbf{r}_{t}(\mathbf{x})/\|\mathbf{r}_{t}(\mathbf{x})\|_2\).

For old classes \(c\in\mathcal{C}_{<t}\), old data are unavailable under the exemplar-free setting. We therefore estimate the raw-space drift using the adversarial anchors \(\mathcal{A}_t^c\) associated with each old class. Specifically, for an anchor \(\mathbf{x}^{adv}\in\mathcal{A}_t^c\), we compute its normalized raw features before and after task-\(t\) training, i.e., \(\bar{\mathbf{r}}^{T}(\mathbf{x}^{adv})\) from the frozen teacher and \(\bar{\mathbf{r}}_{t}(\mathbf{x}^{adv})\) from the updated encoder.
The anchor displacement, reliability score, and normalized weight are defined as
\begin{equation}
\begin{aligned}
\mathbf{d}_t(\mathbf{x}^{adv})
&=
\bar{\mathbf{r}}_{t}(\mathbf{x}^{adv})
-
\bar{\mathbf{r}}^{T}(\mathbf{x}^{adv}), \\
a(\mathbf{x}^{adv},c)
&=
\bar{\mathbf{r}}^{T}(\mathbf{x}^{adv})^\top \boldsymbol{\mu}_{t-1,c}, \\
w(\mathbf{x}^{adv},c)
&=
\frac{a(\mathbf{x}^{adv},c)}
{\sum_{\mathbf{z}\in\mathcal{A}_t^{c}} a(\mathbf{z},c)} .
\end{aligned}
\end{equation}
The class-wise anchor-induced drift and its reliability gate are then estimated by
\begin{equation}
\begin{aligned}
\boldsymbol{\Delta}_{t,c}
&=
\sum_{\mathbf{x}^{adv}\in\mathcal{A}_t^{c}}
w(\mathbf{x}^{adv},c)\mathbf{d}_t(\mathbf{x}^{adv}), \\
g_{t,c}
&=
\frac{1}{|\mathcal{A}_t^{c}|}
\sum_{\mathbf{x}^{adv}\in\mathcal{A}_t^{c}}
\bar{\mathbf{r}}^{T}(\mathbf{x}^{adv})^\top \boldsymbol{\mu}_{t-1,c}.
\end{aligned}
\end{equation}
Finally, the raw visual prototype is transferred as
\begin{equation}
\boldsymbol{\mu}_{t,c}
=
\frac{\boldsymbol{\mu}_{t-1,c}+g_{t,c}\boldsymbol{\Delta}_{t,c}}
{\left\|\boldsymbol{\mu}_{t-1,c}+g_{t,c}\boldsymbol{\Delta}_{t,c}\right\|_2},
\quad \forall c\in\mathcal{C}_{<t}.
\label{eq:old_proto_transfer}
\end{equation}

Unlike existing dual-branch methods that mainly use fixed visual prototypes or auxiliary visual classifiers, we further update old visual prototypes through anchor-induced raw-space drift estimation, so that the visual references remain compatible with the continually adapted visual encoder.

\textit{2) Dual-path Prediction:}
At inference time after task $t$, we combine two complementary decision signals: (i) the cross-modal matching scores produced by CLIP, which inherit the pretrained vision-language semantics, and (ii) the raw-space prototype scores, which capture discriminative visual cues that may not be fully represented by textual embeddings under the modality gap \cite{zhou2025external,li2025bofa}. This dual-path design is particularly suitable for exemplar-free continual learning, where preserving cross-modal semantics is crucial while a lightweight visual reference can further improve robustness. Given a test image $\mathbf{x}$ and a candidate class $c\in\mathcal{C}_{\le t}$, we compute the CLIP branch score and the prototype branch score as
\begin{equation}
s^{\mathrm{clip}}_{t}(\mathbf{x},c)
=
\bar{\mathbf{v}}_{t}(\mathbf{x})^\top \mathbf{u}_{t,c},
\quad
s^{v}_t(\mathbf{x},c)
=
\bar{\mathbf{r}}_{t}(\mathbf{x})^\top \boldsymbol{\mu}_{t,c},
\end{equation}
where $\bar{\mathbf{v}}_{t}(\mathbf{x})=\frac{\mathbf{v}_{t}(\mathbf{x})}{\|\mathbf{v}_{t}(\mathbf{x})\|_2}$ and
$\mathbf{u}_{t,c}=\frac{\mathbf{G}_{t}(\mathbf{pr.}(c))}{\|\mathbf{G}_{t}(\mathbf{pr.}(c))\|_2}$ denote the normalized image and class-text embeddings of the adapted CLIP, and $\bar{\mathbf{r}}_{t}(\mathbf{x})$ and $\boldsymbol{\mu}_{t,c}$ are the normalized raw visual feature and the corresponding class prototype (Sec.~\ref{subsec:prelim}). We then fuse the two branches by a weighted logit ensembling:
\begin{equation}
\ell_t(\mathbf{x},c)
=
s^\mathrm{clip}_{t}(\mathbf{x},c)
+
\beta\, s^{v}_t(\mathbf{x},c),
\qquad c\in\mathcal{C}_{\le t},
\label{eq:dual_path_logits}
\end{equation}
and predict the label by
\begin{equation}
\hat{y}
=
\arg\max_{c\in\mathcal{C}_{\le t}}\ \ell_t(\mathbf{x},c).
\label{eq:dual_path_pred}
\end{equation}

% Preamble (IEEEtran):
% \usepackage{booktabs}
% \usepackage{multirow}

\begin{table*}[t]
\centering
\caption{Comparison with different methods using a CLIP-ViT-B/16 backbone.}
\label{tab:main_cil_results}
\setlength{\tabcolsep}{11pt}
\renewcommand{\arraystretch}{1.0}
\resizebox{\textwidth}{!}{%
\begin{tabular}{l c c cc cc cc cc cc}
\toprule
\multirow{2}{*}{Method} & \multirow{2}{*}{Publication} & \multirow{2}{*}{Replay} &
\multicolumn{2}{c}{CIFAR100$^{(10)}$} &
\multicolumn{2}{c}{ImageNet-R$^{(10)}$} &
\multicolumn{2}{c}{ImageNet-Sub$^{(10)}$} &
\multicolumn{2}{c}{CUB-200$^{(10)}$} &
\multicolumn{2}{c}{UCF$^{(10)}$} \\
\cmidrule(lr){4-5}\cmidrule(lr){6-7}\cmidrule(lr){8-9}\cmidrule(lr){10-11}\cmidrule(lr){12-13}
& & & \textit{Avg} & \textit{Last} & \textit{Avg} & \textit{Last} & \textit{Avg} & \textit{Last} & \textit{Avg} & \textit{Last} & \textit{Avg} & \textit{Last} \\
\midrule
SLCA~\cite{zhang2023slca}                 & \textit{ICCV'23}   &              & 80.5 & 67.6 & 75.9 & 70.4 & 78.6 & 59.9 & 73.3 & 60.4 & --   & --   \\
L2P++~\cite{wang2022learning}             & \textit{CVPR'22}   &              & 81.9 & 73.1 & 81.7 & 76.0 & 80.5 & 67.2 & 71.9 & 63.0 & --   & --   \\
CODA~Prompt~\cite{smith2023coda}          & \textit{CVPR'23}   &              & 77.0 & 62.3 & 78.0 & 67.5 & 64.1 & 34.8 & 66.6 & 50.9 & 87.7 & 82.3 \\
Aper-Adapter~\cite{zhou2025revisiting}    & \textit{IJCV'24}   &              & 75.8 & 65.5 & 78.7 & 71.4 & 85.8 & 76.4 & --   & --   & --   & --   \\
\midrule
\rowcolor[HTML]{EEEEEE}Continual-CLIP~\cite{thengane2022clip} & -- &  & -- & 66.7 & -- & 72.0 & -- & 75.4 & -- & 51.2 & -- & 65.2 \\
CLAP~\cite{jha2024clap4clip}              & \textit{NeurIPS'24}& \checkmark   & 86.1 & 78.2 & 85.8 & 80.0 & 87.8 & 79.2 & \textbf{85.8} & \textbf{80.7} & 91.4 & 86.4 \\
MoE-Adapter~\cite{yu2024boosting}         & \textit{CVPR'24}   &              & 85.4 & 78.4 & 85.3 & 80.8 & 86.4 & 76.7 & --   & --   & --   & --   \\
RAPF~\cite{huang2025class}                & \textit{ECCV'24}   &              & 86.2 & 79.0 & 85.6 & 80.3 & 87.5 & 80.2 & 82.7 & 76.2 & 92.5 & 87.5 \\
LGVLM~\cite{zhang2024continual}                & \textit{TCSVT'24}   &              & 86.3 & 81.1 & 86.8 & 82.1 & -- & -- & -- & -- & -- & -- \\
PROOF~\cite{zhou2025learning}             & \textit{TPAMI'25}  & \checkmark   & 84.9 & 76.3 & 82.8 & 77.1 & 84.7 & 72.5 & 83.1 & 75.5 & 93.6 & 88.9 \\
CLG-CBM \cite{yu2025language} &\textit{CVPR'25} &                     & 84.9 & 76.9 & -- & -- & 86.0 & 78.5 & 82.9& 77.8 & --& -- \\
DMNSP \cite{kang2025dynamic}  &\textit{ICCV'25} &                     & 87.1 & 79.9 & 82.7 & 76.2& 85.0 & 76.1 & -- & -- & --& -- \\
ENGINE~\cite{zhou2025external}            & \textit{ICCV'25}   &              & 82.1 & 73.1 & 84.4 & 77.0 & --   & --   & 83.9 & 76.2 & 95.0 & 90.1 \\
MG-CLIP~\cite{huang2025mind}              & \textit{ICCV'25}   &              & 87.0 & 80.6 & 87.6 & 82.7 & 87.3 & 78.4 & 80.6 & 72.0 & --   & --   \\
DesCLIP~\cite{he2025desclip}              & \textit{TMM'26}   &              & 85.6 & 78.7 & 85.1 & 78.7 & 86.4 & 77.8 & 80.7 & 72.2 & --   & --   \\
\midrule
\textbf{SeGP-CL}-\textit{OnlyCLIP}        &       &              & 88.2 & 83.2 & 87.4 & 83.9 & 88.7 & 78.8 & 83.0 & 74.8 & 91.4 & 88.9 \\
\rowcolor[HTML]{D7D2F5}\textbf{SeGP-CL} (Ours) &  &          & \textbf{89.8} & \textbf{84.6} & \textbf{88.9} & \textbf{84.8} & \textbf{89.9} & \textbf{80.5} & 85.4 & 80.1 & \textbf{95.9} & \textbf{92.8} \\
\bottomrule
\end{tabular}
}
\end{table*}

\begin{figure*}[t]
  \centering
  \includegraphics[width=0.85\linewidth]{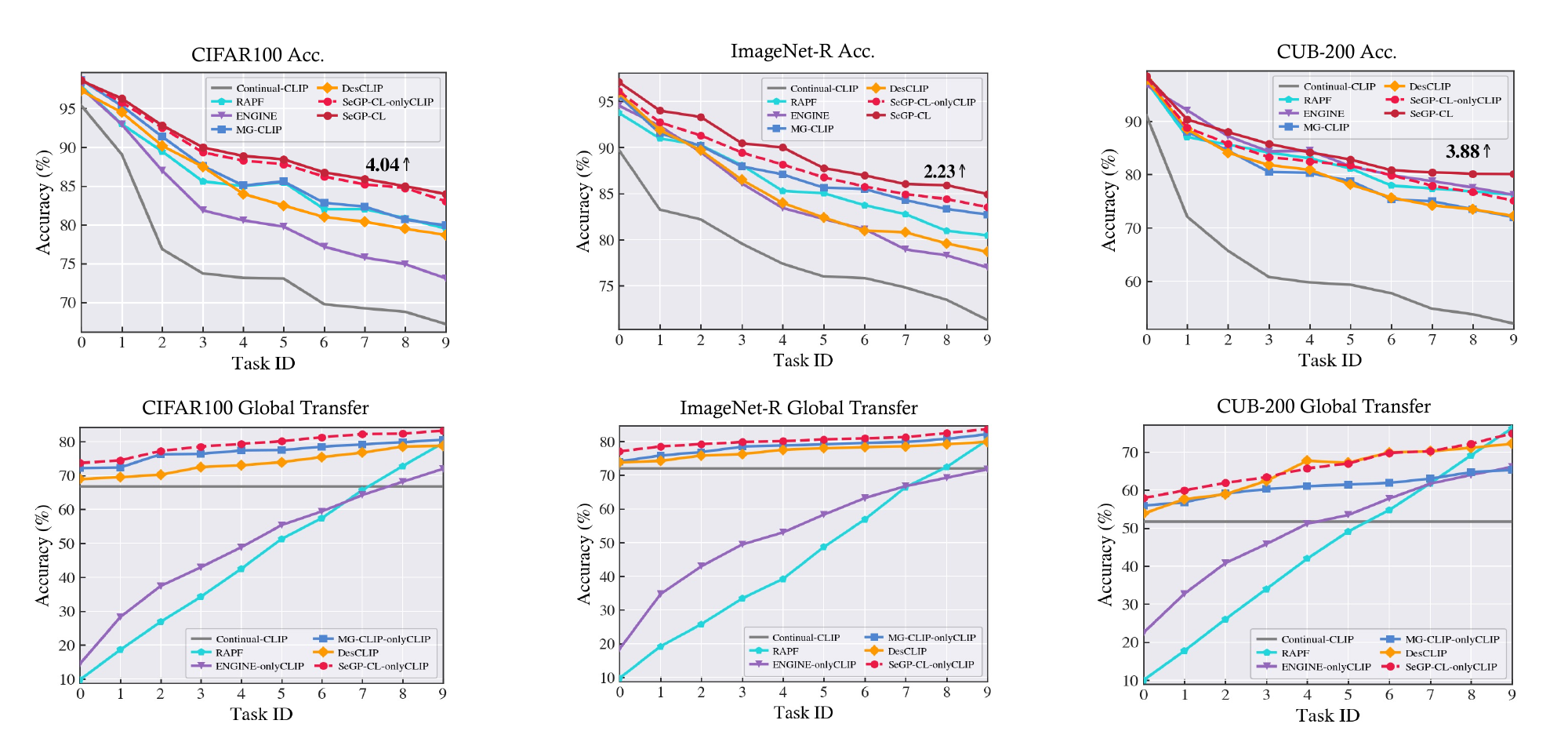}
  \vspace{-1em}
  \caption{Comparison with state-of-the-art CL methods in terms of per-task accuracy and global transfer. All results are achieved on the same CLIP ViT-B/16 backbone.  }
  \label{fig:acc-forget-curve}
\end{figure*}

\section{Experiments}
\subsection{Setup}

\textit{1) Benchmark: } Our experiments for VLM-based continual learning are conducted on the common CL benchmark, including CIFAR100 \cite{krizhevsky2009learning}, ImageNet-Sub \cite{deng2009imagenet}, CUB-200 \cite{wah2011caltech}, ImageNet-R \cite{hendrycks2021imagenet}, and UCF101 \cite{soomro2012ucf101}.

\textit{2) Metrics:} To evaluate VLM-based continual learning, we report two primary metrics: \textit{Last} and \textit{Avg}. \textit{Last} denotes the average classification accuracy over all seen classes after training on the final task, while \textit{Avg} is the mean incremental accuracy averaged over tasks. To further characterize knowledge transfer and retention, we also report \textit{Forward Transfer} (FWT), \textit{Backward Transfer} (BWT), and \textit{Forgetting}. Following the standard protocol, let $R_{i,j}$ be the test accuracy on task $j$ after completing training on task $i$. Then
\begin{equation}
\mathrm{BWT}
=
\frac{1}{t-1}\sum_{j=1}^{t-1}\big(R_{t,j}-R_{j,j}\big),
\quad
\mathrm{FWT}
=
\frac{1}{t-1}\sum_{j=2}^{t} R_{j-1,j},
\end{equation}
and the \textit{Forgetting} is computed as
\begin{equation}
\mathrm{Forgetting}
=
\frac{1}{t-1}\sum_{j=1}^{t-1}\Big(\max_{i\le t} R_{i,j}-R_{t,j}\Big).
\end{equation}

\textit{3) Competitors:} We compare our method against baseline and state-of-the-art methods for continual learning. These include \textbf{(i)Vision-based CL approaches:} L2P++ \cite{wang2022learning}, CODA Prompt \cite{smith2023coda} and SLCA \cite{zhang2023slca}; \textbf{(ii)VLM-based approaches:} ContinualCLIP \cite{thengane2022clip} (zero-shot baseline), PROOF \cite{zhou2025learning}, CLAP \cite{jha2024clap4clip}, LGVLM \cite{zhang2024continual}, MoE-Adapters \cite{yu2024boosting}, RAPF \cite{huang2025class}, CLG-CBM \cite{yu2025language}, DMNSP \cite{kang2025dynamic}, ENGINE \cite{zhou2025external}, DesCLIP \cite{he2025desclip}, and MG-CLIP \cite{huang2025mind}.

%\textit{4) Implementation Details:} All experiments are conducted on two NVIDIA GeForce RTX 4090 GPUs. We adopt the pre-trained CLIP \cite{radford2021learning} ViT-B/16 from OpenAI as the backbone. To ensure a fair comparison, all methods are built upon the same pre-trained backbones.

\textit{4) Implementation Details:}

All experiments are conducted on two NVIDIA GeForce RTX 4090 GPUs. We adopt the pre-trained OpenAI CLIP ViT-B/16~\cite{radford2021learning} as the default backbone for our method and all VLM-based competitors. To ensure a fair comparison, all reproduced methods are evaluated under the same CL protocol, including the same task order, class-to-task split, training schedule, and evaluation metrics. For methods whose original papers report results with different CLIP variants or pre-training sources, such as CLIP ViT-L/14 or OpenCLIP \cite{cherti2023reproducible}, we use their ViT-B/16 results when available or reproduce them with the official implementation under the unified OpenAI CLIP ViT-B/16 backbone.

The model is trained for 10 epochs on each incremental task. We use stochastic gradient descent (SGD) \cite{bottou2010large} with a cosine learning-rate decay schedule, a batch size of 128, and an initial learning rate of 0.001. Following \cite{huang2025mind}, we insert LoRA \cite{hu2022lora}(rank = 32) into both the visual and textual encoders of CLIP by re-parameterizing the Transformer attention projection matrices \textit{K} and \textit{V}, and the linear layers in the Feedforward Network (\textit{FFN}). During continual adaptation, we freeze the LoRA down-projection matrices (\textit{A}) and only update the up-projection matrices (\textit{B}). For the anchor construction process, we set \(\lambda_p=0.5\) to form the corrected objective \(\mathcal{L}^{\prime}_{\mathrm{adv}}\), use an \(\ell_{\infty}\) perturbation radius of \(\epsilon=4/255\), and run \(K_{\mathrm{adv}}=10\) DPGD iterations with a step size of \(\gamma=1.5\times10^{-3}\). For each old prototype, we select $K_{\mathrm{seed}}=5$ seed samples for attack and use the 5 anchors for ACGD. During training, we set the anchor batch size to 32 for distillation. For the distillation temperatures, we use $\tau_A=20$ and $\tau_T=0.05$. The loss weights are set to $\lambda_{\mathrm{ACGD}}=5$ and $\lambda_{\mathrm{GR}}=1$. For TSGR, the $k$ in the $k$-NN subgraph is set to 10. For logit ensembling at inference, we set the visual-branch coefficient to $\beta=0.5$.

\begin{table}[!t]
\centering
\setlength{\tabcolsep}{8pt}
\renewcommand{\arraystretch}{1.0}
\caption{Comparison on forward transfer (FWT), backward transfer (BWT) and forgetting (F) of the global visual-text matching.}
\label{tab:compare_fwt_bwt_forgetting}
\resizebox{0.5\textwidth}{!}{
\begin{tabular}{lcccccc}
\toprule[1.pt]
\multirow{2}{*}{\textbf{Methods}(\textit{On'l'y})} &
\multicolumn{3}{c}{\textbf{CIFAR100}} &
\multicolumn{3}{c}{\textbf{ImageNet-R}} \\
\cmidrule(lr){2-4}\cmidrule(lr){5-7}
& FWT$\uparrow$ & BWT$\uparrow$ & F$\downarrow$
& FWT$\uparrow$ & BWT$\uparrow$ & F$\downarrow$ \\
\midrule
ENGINE   & \textcolor{red}{0} & -10.9 & 7.9 & \textcolor{red}{0} & -7.0 & 7.1 \\
RAPF     & \textcolor{red}{0} & -8.9  & 8.5 & \textcolor{red}{0} & -6.1 & 6.6 \\
DesCLIP  & 68.7 & -2.1  & 6.5 & 74.9 & -1.0 & 1.7 \\
MG-CLIP  & 70.2 & -3.9  & 4.9 & 76.2 & -0.01 & 0.88\\
\rowcolor[HTML]{D7D2F5}
\textbf{SeGP-CL} & \textbf{72.3} & \textbf{-0.43} & \textbf{0.9} & \textbf{77.0} & \textbf{+0.03} & \textbf{0.32} \\
\bottomrule[1.pt]
\end{tabular}
}
\end{table}

\subsection{Comparison with State-of-the-art Methods}
\label{sec:Comparison with State-of-the-art Methods}
Table~\ref{tab:main_cil_results} summarizes the results on five benchmarks in terms of \textit{Avg} and \textit{Last}. 
Compared with both vision-only continual learning baselines SLCA~\cite{zhang2023slca}, L2P++~\cite{wang2022learning} and VLM-oriented approaches RAPF~\cite{huang2025class}, ENGINE~\cite{zhou2025external}, and MG-CLIP~\cite{huang2025mind}), our method consistently achieves the best overall performance without any data replay.
In particular, on CIFAR100, SeGP-CL improves the \textit{Last} accuracy from 80.6 (MG-CLIP) to 84.6 (+4.0), and on CUB-200 it boosts \textit{Last} from 76.2 (RAPF) to 80.1 (+3.9). 
Notably, even under the \textit{OnlyCLIP} setting (relying purely on CLIP vision-text matching without additional vision-branch predictors), our method remains highly competitive and still surpasses prior methods that typically depend on extra visual-branch components (e.g. ENGINE, MG-CLIP). Beyond these, Fig.~\ref{fig:acc-forget-curve}(Upper) provides a per-task comparison across the entire incremental process, where our approach maintains the highest seen-task average accuracy and exhibits the smallest degradation on previous tasks, suggesting substantially reduced forgetting.

Moreover, since CLIP is endowed with strong zero-shot ability, we explicitly assess whether such inherent visual-textual cognition can be progressively accumulated during downstream incremental adaptation. Specifically, we adopt a global evaluation protocol: after each incremental stage, we evaluate all test samples from all tasks, and predict each sample against the entire class set solely via CLIP's image-text matching, without any auxiliary vision-side classifier. As shown in Fig.~\ref{fig:acc-forget-curve}(Bottom), under this global transfer setting, our SeGP-CL exhibits stable and monotonic improvement. In contrast, recent approaches such as RAPF and ENGINE, which rely on task-specific components, may perturb the shared vision-language embedding space and thereby undermine global image-text matching, resulting in weakened forward transfer. In other words, our method truly incrementally accumulates downstream knowledge from the pretrained state through controlled adaptation, rather than merely re-organizing.
These observations are further supported by the metrics in Table~\ref{tab:compare_fwt_bwt_forgetting}, where our method achieves the strongest \textit{Forward Transfer} (FWT), \textit{Backward Transfer} (BWT), and the lowest \textit{Forgetting} (F), demonstrating the effectiveness of preserving cross-modal semantic geometry.

\begin{table}[t]
\centering
\caption{Comparison of distillation schemes on CIFAR100 under different data-source and sample-budget settings. For the Naive baseline, values in parentheses denote the absolute \(\overline{A}_{\mathrm{new}}\) and \textit{Last}; other rows report changes relative to this baseline.}
\label{tab:distill_data_ablation}
\setlength{\tabcolsep}{4.2pt}
\renewcommand{\arraystretch}{1.05}
\resizebox{\columnwidth}{!}{%
\begin{tabular}{lcccc}
\toprule[1.pt]
\textbf{Constraint} &
\textbf{Distillation Source} &
\textbf{Sample Budget} &
\(\Delta \overline{A}_{\mathrm{new}}\uparrow\) &
\(\Delta\)\textit{Last}\(\uparrow\) \\
\midrule

\rowcolor[HTML]{EEEEEE}
Naive baseline & -- & -- & (89.8) & (77.0) \\

\midrule
\multicolumn{5}{l}{\textit{Distillation on full new-task data}} \\
SGCL~\cite{yu2024exploiting} & New-task data & \(N_t\) & -3.2 & +2.5 \\
CGD & New-task data & \(N_t\) & -2.8 & +1.5 \\

\midrule
\multicolumn{5}{l}{\textit{Matched small-budget distillation sources}} \\
SGCL~\cite{yu2024exploiting} &
Random new-task data &
\(N_t \rightarrow M_t\) &
-3.1\({\scriptstyle\pm 0.3}\) & +2.4\({\scriptstyle\pm 0.2}\) \\

CGD &
Random new-task data &
\(N_t \rightarrow M_t\) &
-2.7\({\scriptstyle\pm 0.4}\) & +1.5\({\scriptstyle\pm 0.2}\) \\
CGD & Synthetic old data~\cite{wu2025synthetic} & \(M_t\) & -3.8 & \textcolor{red}{-3.5} \\
CGD & Selected seeds & \(N_t \rightarrow M_t\) & -1.4 & +3.1 \\
\rowcolor[HTML]{D7D2F5}
CGD & Adv. anchors & \(M_t\) & \textbf{-0.5} & \textbf{+5.8} \\
\textcolor{gray}{CGD} &
\textcolor{gray}{Real old data} &
\textcolor{gray}{\(M_t\)} &
\textcolor{gray}{-0.8} &
\textcolor{gray}{\textbf{+7.0}} \\

\midrule
\multicolumn{5}{l}{\textit{External general sources}} \\
ZSCL~\cite{zheng2023preventing} & CC Val. Ref.~\cite{changpinyo2021conceptual} & 28K & -2.3 & +1.9 \\
DualTeacher~\cite{yu2025select} & ImageNet Ref.~\cite{deng2009imagenet} & 100K & -2.0 & +1.5 \\

\bottomrule[1.pt]
\end{tabular}
}
\end{table}

\begin{table}[t]
\centering
\setlength{\tabcolsep}{8pt}
\renewcommand{\arraystretch}{1.0}
\caption{Ablation on key components. `PT' denotes using Prototype Transfer and `V.' denotes using an additional visual branch for prediction.}
\label{tab:ablate_components}
\resizebox{\columnwidth}{!}{%
\begin{tabular}{cccccccc}
\toprule[1.pt]
\multicolumn{4}{c}{\textbf{Components}} &
\multicolumn{2}{c}{\textbf{CIFAR100}} &
\multicolumn{2}{c}{\textbf{UCF}} \\
\cmidrule(lr){5-6}\cmidrule(lr){7-8}
w/$~\mathcal{L}_{\mathrm{ACGD}}$ & w/$~\mathcal{L}_{\mathrm{GR}}$ & w/ PT & w/ V. &
\textit{Last} & F$\downarrow$ &
\textit{Last} & F$\downarrow$ \\
\midrule
\rowcolor[HTML]{EEEEEE}-- & -- & -- & -- & 77.0 & 10.9 & 84.4 & 10.7 \\
\checkmark &  &  &  & 81.7 & 7.0 & 88.2 & 6.6 \\
\checkmark & \checkmark &  &  & 82.8 & 5.9 & 88.0 & 6.9 \\
\checkmark & \checkmark & \checkmark &  & 83.2 & \textbf{5.3} & 88.9 & \textbf{6.3} \\
\checkmark & \checkmark & \checkmark & \checkmark & \textbf{84.6} & 5.7 & \textbf{92.8} & 6.8 \\
\bottomrule[1.pt]
\end{tabular}
}
\end{table}

\begin{table}[t]
\centering
\caption{Sensitivity analysis of the visual-target correction weight \(\lambda_p\) in DPGD. The reported metric is \textit{Last} accuracy.}
\label{tab:lambda_p_sensitivity}
\setlength{\tabcolsep}{3.2pt}
\renewcommand{\arraystretch}{1.}
\resizebox{0.75\columnwidth}{!}{%
\begin{tabular}{lccccccc}
\toprule[1.pt]
\textbf{Dataset} &
\makecell{\(\lambda_p=0\)\\w/o \(\mathcal{L}_{\mathrm{v\text{-}adv}}\)} &
0.2 & 0.3 & 0.4 & 0.5 & 0.6 & 0.7 \\
\midrule
CIFAR100 & \color{red}{82.7} & 83.8 & 84.2 & \textbf{84.6} & \textbf{84.6} & 84.4 & 84.2 \\
CUB-200  & \color{red}{78.1} & 79.6 & 79.9 & 79.9 & 80.1 & \textbf{80.2} & 79.8 \\
\bottomrule[1.pt]
\end{tabular}
}
\end{table}

\begin{table}[t]
\centering
\caption{Sensitivity analysis of the number of seed samples \(K_{\mathrm{seed}}\) on ImageNet-R. The reported metric is \textit{Last} accuracy.}
\label{tab:kseed_sensitivity}
\setlength{\tabcolsep}{3.6pt}
\renewcommand{\arraystretch}{1.0}
\resizebox{\columnwidth}{!}{%
\begin{tabular}{lcccccc}
\toprule[1.pt]
\multirow{2}{*}{\textbf{Selection}} &
\multicolumn{6}{c}{\(K_{\mathrm{seed}}\)} \\
&
Baseline (0) & 1 & 3 & 5 & 10 & 20 \\
\midrule
Random &
\cellcolor[HTML]{EEEEEE}80.1 &
81.4{\scriptsize\(\pm1.7\)} &
81.6{\scriptsize\(\pm1.2\)} &
82.0{\scriptsize\(\pm0.8\)} &
81.7{\scriptsize\(\pm0.6\)} &
82.2{\scriptsize\(\pm0.5\)} \\
\(Q(\mathbf{x},c)\) &
\cellcolor[HTML]{EEEEEE}80.1 &
82.1 & 83.6 & 84.8 & 85.2 & \textbf{85.4} \\
\bottomrule[1.pt]
\end{tabular}
}
\end{table}

\begin{figure}[t]
  \centering
  \includegraphics[width=\linewidth]{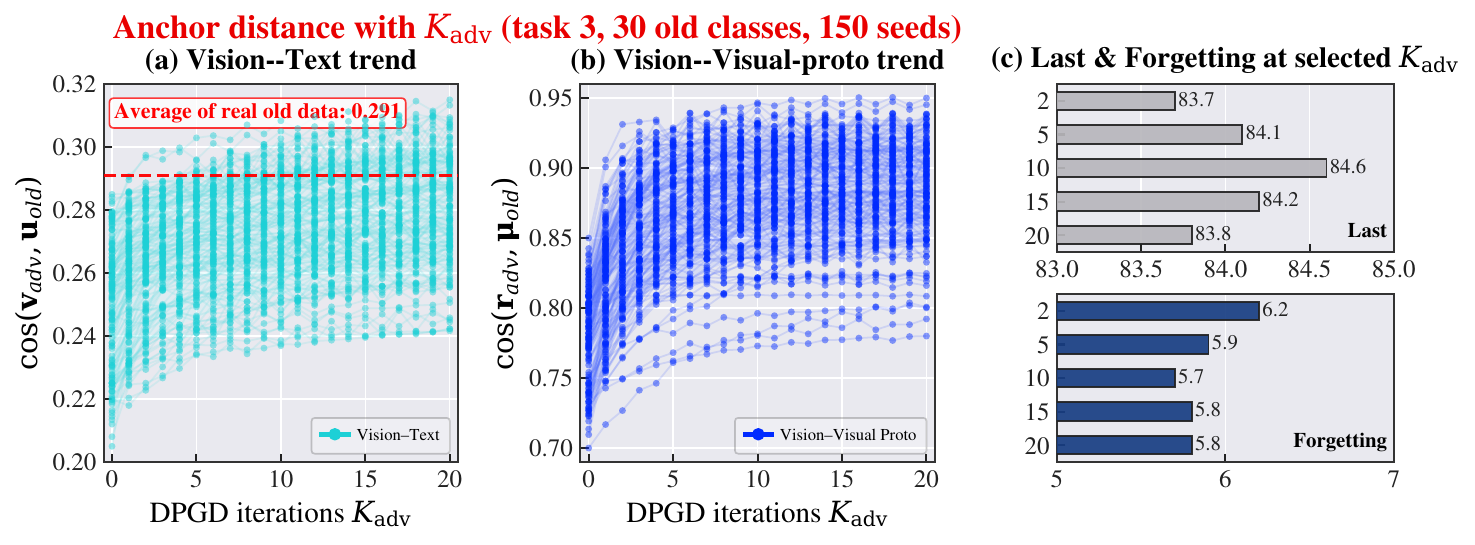}
  \vspace{-2em}
  \caption{Effect of adversarial iterations $K_{\mathrm{adv}}$. We visualize the distance evolution of anchor construction across DPGD iterations in (a) CLIP embedding space and (b) raw visual feature space, and report the corresponding continual learning performance in (c).}
  \label{fig:k_adv}
\end{figure}

\begin{figure}[t]
  \centering
  \includegraphics[width=0.9\linewidth]{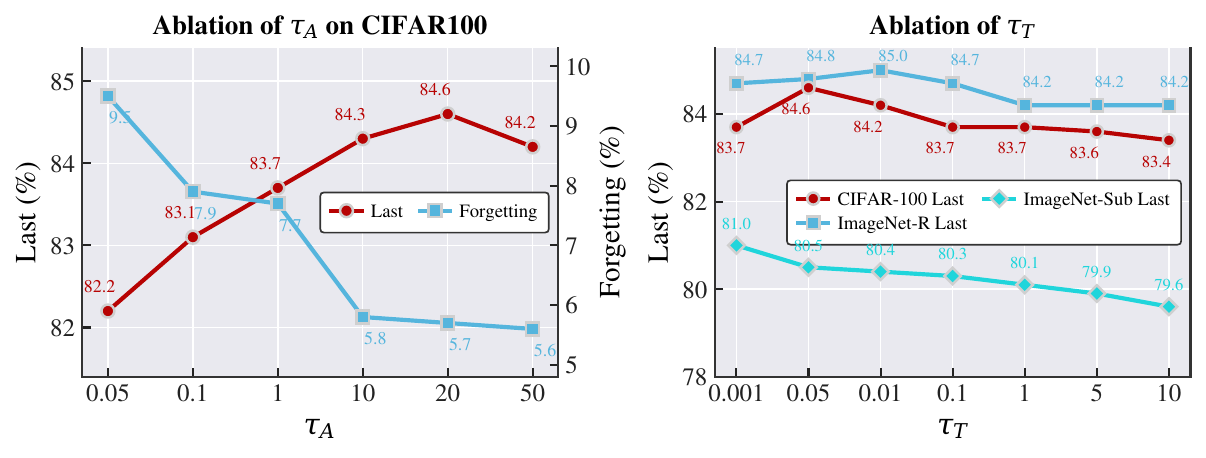}
  \vspace{-1em}
  \caption{Ablation of temperature $\tau_A$ and $\tau_T$.}
  \label{fig:ablation_temp}
\end{figure}

\begin{figure}[t]
  \centering
  \includegraphics[width=0.9\linewidth]{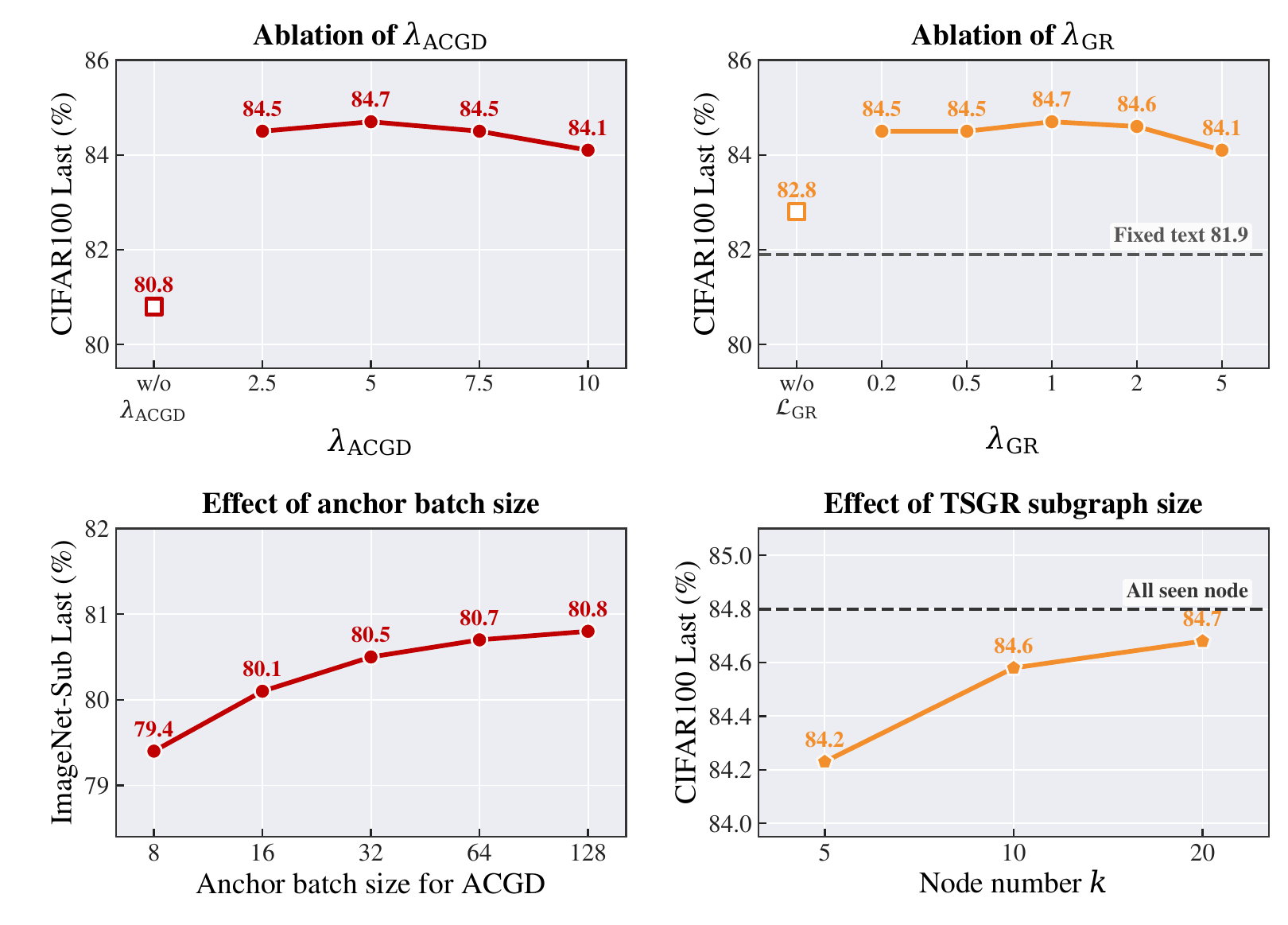}
  \vspace{-1em}
  \caption{Ablation of \(\lambda_{\mathrm{ACGD}}\), \(\lambda_{\mathrm{GR}}\), anchor batch size, and TSGR subgraph size.}
  \label{fig:adv+subgraph}
\end{figure}

\begin{figure}[t]
  \centering
  \includegraphics[width=0.85\linewidth]{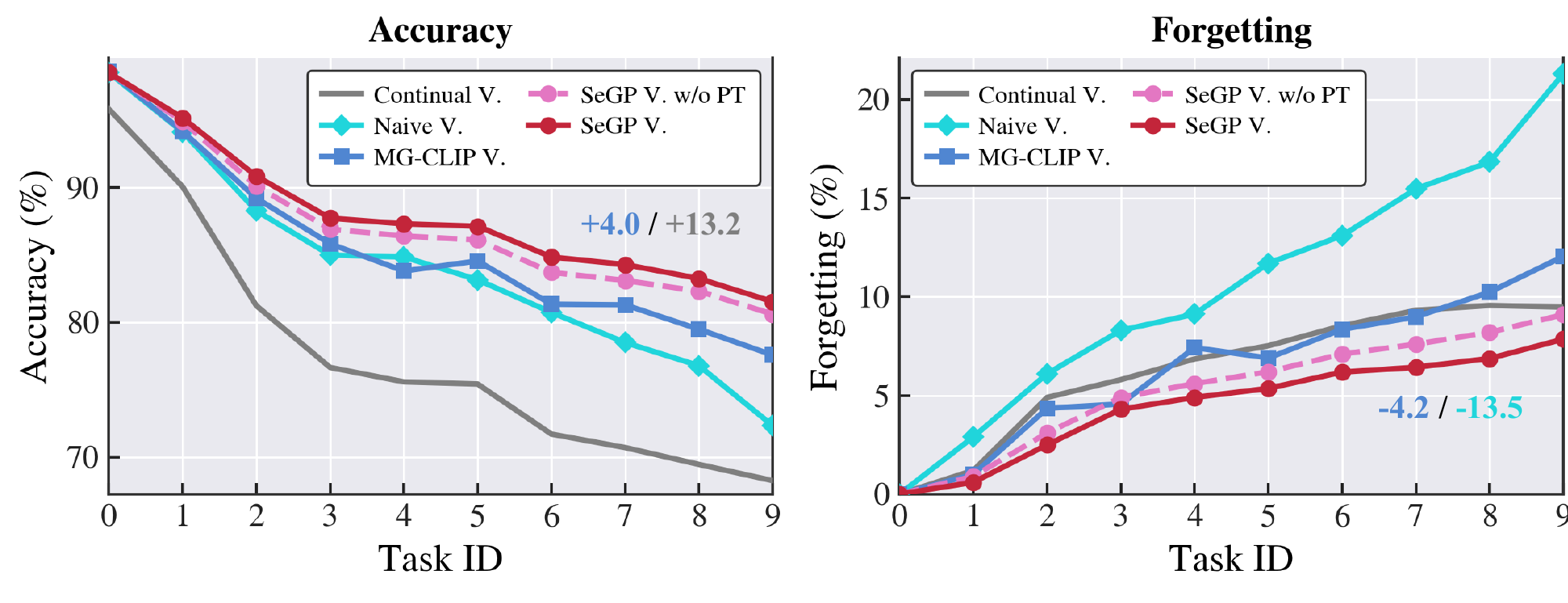}
  \vspace{-1em}
  \caption{Evaluation of cross-modal transfer robustness on CIFAR100. We annotate the gains in \textit{Last} accuracy and \textit{Forgetting} relative to the baselines.}
  \label{fig:V.}
\end{figure}

% (need packages) \usepackage{booktabs}
\begin{table}[t]
\centering
\setlength{\tabcolsep}{5pt}
\renewcommand{\arraystretch}{1.0}
\caption{Cross-scenario robustness evaluation.}
\label{tab:zs_transfer}
\resizebox{\columnwidth}{!}{%
\begin{tabular}{l cc c c}
\toprule[1.pt]
\textbf{Method} & \textbf{CIFAR-\textit{Last}}&\textbf{Food101} & \textbf{Oxford-Pets} & \textbf{ImageNet-full(1K)} \\
\midrule
\rowcolor[HTML]{EEEEEE}\textit{zero-shot} CLIP  & 66.7& 85.1 & 87.6 & 64.4 \\
PROOF   & 76.3 & \textcolor{red}{9.8}  & \textcolor{red}{22.8} & \textcolor{red}{11.2}  \\
RAPF  & 79.0 & \textcolor{red}{17.2} & \textcolor{red}{28.6} & \textcolor{red}{15.2}  \\
MoE-Adapter &78.4 & 82.9 & 84.1 & 66.0  \\
SPU &77.0& 84.3 & 85.9 & 62.4 \\
MG-CLIP & 80.6 &\textbf{85.7} & \textbf{88.2} & 67.3 \\
\midrule
\textbf{SeGP-CL} w/o TSGR & 84.2 & 82.1 & 85.2 & 62.4 \\
\rowcolor[HTML]{D7D2F5}\textbf{SeGP-CL} &\textbf{84.6}& 84.5 & 87.7 & \textbf{67.6} \\
\bottomrule[1.pt]
\end{tabular}
}
\end{table}

\begin{table}[t]
\centering
\setlength{\tabcolsep}{5pt}
\renewcommand{\arraystretch}{1.}
\caption{Comparison on trainable parameters and time cost. \textit{K,V} represents the \textit{key/value} attention projection weights. \underline{\textbf{Underlined}} values indicate the baseline average time.}
\label{tab:lora_rank_inject}
\resizebox{\columnwidth}{!}{%
\begin{tabular}{c l c c c}
\toprule[1.pt] 
\multicolumn{2}{c}{\textbf{Setting}} & \textbf{CIFAR-\textit{Last}} & \textbf{\#Trainable Params} & \textbf{Avg Time}(ms) / Iter. \\
\midrule
\multirow{3}{*}{\shortstack{Ours\\(rank=8)}}  & \textit{K,V}       & 83.1 & 0.25M & \underline{330}+53 \\
                                            & \textit{FFN}       & 82.3 & 0.61M & \underline{401}+69 \\
                                            & \textit{K,V+FFN}   & 83.4 & 0.86M & \underline{410}+74 \\
\midrule
\multirow{3}{*}{\shortstack{Ours\\(rank=32)}} & \textit{K,V}       & 83.9 & 0.98M & \underline{339}+55 \\
                                            & \textit{FFN}       & 83.1 & 2.46M & \underline{419}+68 \\
                                            & \textit{K,V+FFN}   & 84.6 & 3.44M & \underline{424}+79 \\
\midrule
\multicolumn{2}{l}{MG-CLIP (rank=8)}  & 80.6 & 0.54M & 333 \\
\multicolumn{2}{l}{MG-CLIP (rank=32)} & 80.8 & 2.02M & 343 \\
\multicolumn{2}{l}{ZSCL (full FT)}      & 74.2 & 149.6M & 4205 \\
\multicolumn{2}{l}{MoE-Adapter}       & 78.4 & 13.35M & 1679 \\
\multicolumn{2}{l}{DMNSP}              & 79.9 & 7.8M & -- \\
\multicolumn{2}{l}{CLAP}              & 78.2 & 9.45M & -- \\
\bottomrule[1.pt]
\end{tabular}
}
\end{table}

\subsection{Comparison with Distillation Schemes}
{
Table~\ref{tab:distill_data_ablation} compares different distillation schemes and data sources. We report \(\Delta \overline{A}_{\mathrm{new}}\), the change of the average newly learned task accuracy relative to the Naive LoRA baseline, and \(\Delta\textit{Last}\), the final accuracy gain over the baseline. The comparison is organized into full new-task data \(N_t\), matched small-budget sources \(M_t=K_{\mathrm{seed}}|\mathcal{C}_{<t}|\), and external general sources. Here, \(N_t\rightarrow M_t\) means that the source is reduced from all current-task samples to the matched budget by random sampling or seed selection.

The matched-budget results show that the distillation source is more critical than the number of new-task samples. When SGCL \cite{yu2024exploiting} and CGD are reduced from \(N_t\) to \(M_t\), their \(\Delta\textit{Last}\) remains almost unchanged, i.e., +2.5 to +2.4 for SGCL and +1.5 to +1.5 for CGD, while both still suffer from clear new-task accuracy drops. This suggests that simply increasing raw new-task samples does not resolve the conflict between old-class distillation and new-class learning. Moreover, synthetic old data generated by Stable Diffusion~\cite{rombach2022high} from class-name prompts performs poorly under the same budget, indicating that prompt-generated images may deviate from the original domain of old-task knowledge and thus provide noisy distillation signals. In contrast, selected seeds already outperform random new-task samples, improving \(\Delta\textit{Last}\) to +3.1 with a smaller plasticity cost. This confirms that old-related samples are more suitable for preserving established cross-modal geometry. By further moving these seeds toward old semantic neighborhoods, adversarial anchors achieve the best exemplar-free trade-off, with only -0.5 in \(\Delta \overline{A}_{\mathrm{new}}\) and +5.8 in \(\Delta\textit{Last}\). Thus, the gain mainly comes from a targeted boundary-aware distillation source rather than from a larger sample budget. Moreover, our method approaches the gain obtained by distillation with real old data, while showing better new-task plasticity. Although large-scale external data for preserving generic knowledge, such as those used by ZSCL \cite{zheng2023preventing} and DualTeacher \cite{yu2025select}, is effective over the baseline, our method achieves better performance with a much smaller budget and without introducing external data.
}

\subsection{Ablation Study}
\label{sec:ablation}
\textit{1) Component Effectiveness:} We analyze how each key component of our proposed SeGP-CL contributes to the continual learning performance, as summarized in Table~\ref{tab:ablate_components}. We observe that ACGD substantially alleviates forgetting of previously learned knowledge. Moreover, TSGR and Prototype Transfer (PT) further stabilize the semantic geometry of the VLM, leading to improved knowledge retention and overall performance.

\textit{2) Anchor Construction Analysis:} We further investigate the sensitivity of our approach to the anchor construction process on CIFAR-100, as shown in Fig.~\ref{fig:k_adv}.
By running DPGD for multiple iterations, we progressively construct anchor samples that move within the local neighborhood and gradually approach the old-class semantic prototypes from two references: the text embedding and the raw-visual prototype.
Notably, due to the modality gap \cite{liang2022mind}, PGD guided by the raw-visual prototype typically converges faster, whereas the text-prototype branch exhibits more pronounced oscillations during optimization.
Our results also suggest that more iterations are not always better: making anchors arbitrarily close to the old prototypes does not necessarily yield larger distillation gains.
In particular, we observe the best trade-off around $K_{\mathrm{adv}}=10$, which achieves the highest \textit{Last} accuracy and the lowest Forgetting.
When further increasing $K_{\mathrm{adv}}$, both \textit{Last} and \textit{Forgetting} deteriorate.
This phenomenon highlights the necessity of protecting cross-modal geometry specifically in semantically vulnerable neighborhoods, rather than enforcing overly aggressive prototype attraction.

We further analyze anchor-construction hyperparameters $\lambda_p$ and $K_{\mathrm{seed}}$ in Tables~\ref{tab:lambda_p_sensitivity} and~\ref{tab:kseed_sensitivity}. Table~\ref{tab:lambda_p_sensitivity} shows that removing the visual-prototype correction, i.e., \(\lambda_p=0\), clearly reduces the \textit{Last} accuracy on both CIFAR100 and CUB-200. This indicates that text-only adversarial guidance is insufficient under the modality gap, while a moderate visual correction helps anchors better approximate old semantic neighborhoods. The performance remains stable for a range of \(\lambda_p\) values, and the default \(\lambda_p=0.5\) provides a robust trade-off across datasets. Table~\ref{tab:kseed_sensitivity} further shows that the quality of selected seeds is more important than simply increasing their number. Random selection only brings limited improvements even with more seeds, whereas \(Q(\mathbf{x},c)\)-based selection consistently achieves stronger performance. Although increasing \(K_{\mathrm{seed}}\) can further improve \textit{Last} accuracy, \(K_{\mathrm{seed}}=5\) already captures most of the gain while keeping the anchor budget compact.

\begin{figure}[t]
    \centering
    \includegraphics[width=0.9\linewidth]{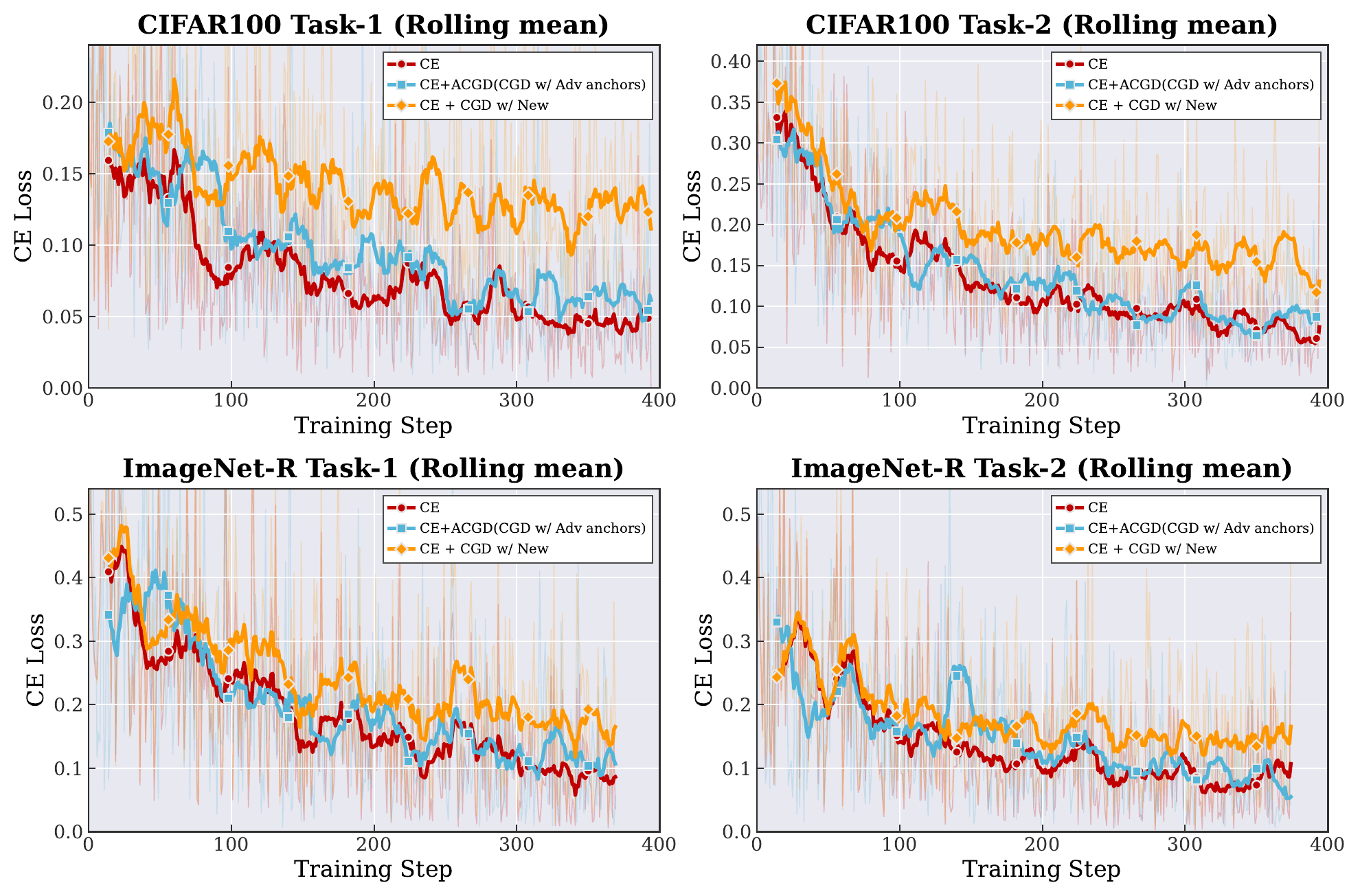}
    \vspace{-1em}
    \caption{CE-loss curves for new-task plasticity analysis.}
    \label{fig:plasticity_ce_loss}
\end{figure}

\textit{3) Analysis of Distillation Configurations:} We further examine the distillation configurations used in continual learning. As shown in Fig.~\ref{fig:ablation_temp}, we vary the distillation temperatures $\tau_A$ and $\tau_T$ to study their effects. We find that ACGD benefits from a relatively high temperature. In particular, $\tau_A=20$ achieves the highest \textit{Last} accuracy and a low forgetting value, providing a favorable stability-plasticity trade-off. This indicates that anchor-guided distillation is more effective when it prioritizes preserving the global cross-modal geometry (larger $\tau_A$) instead of over-emphasizing overly sharp, local relations (smaller $\tau_A$). In contrast, TSGR favors a lower temperature $\tau_T$, highlighting the importance of maintaining a compact and semantically relevant local relational subgraph in the text space during continual updates.

In addition, we examine the sensitivity of the two geometry-preserving loss weights, the anchor batch size, and the TSGR subgraph size in Fig.~\ref{fig:adv+subgraph}. Removing ACGD or TSGR leads to clear performance drops, while moderate values of \(\lambda_{\mathrm{ACGD}}\) and \(\lambda_{\mathrm{GR}}\) consistently maintain strong results. This shows that both anchor-guided cross-modal distillation and text semantic-geometry regularization are necessary, and that SeGP-CL is not sensitive to a narrow loss-weight choice. For the efficiency-related configurations, ACGD remains effective with a small anchor batch size, and larger batches provide only gradual improvements. For TSGR, a compact \(k\)-NN text subgraph already approaches the result of using all seen nodes, indicating that local semantic-geometry regularization is sufficient and more efficient than a global text-graph constraint. We find that ACGD remains robust even with a small anchor batch size (e.g., 16 or 32), while larger batches further strengthen cross-modal geometry preservation. For TSGR, enforcing the constraint on a compact $k$-NN neighborhood around each new-class text embedding is sufficient to regularize semantic relations effectively. In particular, using $k{=}10$ neighbors per new-class node achieves performance close to a global constraint over all seen nodes, while substantially reducing the computational overhead.

\subsection{Robustness Evaluation}
1) \textit{Cross-scenario Robustness Evaluation:} Preserving the generic cross-modal zero-shot capability of VLMs while continually adapting to downstream tasks is crucial \cite{zhang2024overcoming,huang2025mind,zheng2023preventing}. Following~\cite{huang2025mind}, we evaluate the zero-shot transferability of CLIP after continual tuning on several cross-domain benchmarks. Specifically, we include Food-101~\cite{bossard2014food}, Oxford-Pets~\cite{parkhi2012cats}, and ImageNet-full (1K) \cite{deng2009imagenet} to assess the intrinsic cross-modal ability of the model after completing all 10 CL stages on CIFAR100. As shown in Table~\ref{tab:zs_transfer}, while achieving the best CL performance on CIFAR100, our method still maintains zero-shot accuracy comparable to, and even slightly higher than, the original \textit{zero-shot} CLIP on Food-101, Oxford-Pets, and ImageNet-full. This suggests that our adaptation preserves the pretrained cross-modal alignment structure and enables downstream knowledge to be accumulated on top of CLIP, instead of being re-organized by task-specific auxiliary components. In contrast, methods that introduce incremental-task-specific modules (e.g., PROOF, RAPF, and ENGINE) tend to re-route predictions through task-dependent heads, which can distort the shared vision-language embedding space and weaken the global image-text matching capability, leading to notably degraded cross-scenario robustness.

\textit{2) Cross-modal Transfer Robustness.}
While SeGP-CL is primarily designed to preserve underlying cross-modal and text-semantic geometry, we find that this stabilization can transfer back to the visual modality and improve purely visual-branch continual adaptation (predict as Eq. \ref{eq:V. pred}) as well. Intuitively, by anchoring the student updates with ACGD and maintaining a stable textual semantic reference frame via TSGR, the visual encoder is guided to update along directions that remain compatible with the pretrained vision-language structure, which in turn yields more stable and discriminative raw visual representations. Together with the after-training prototype transfer (PT), these cross-modal constraints provide an implicit but effective regularization for the visual branch. We refer to this phenomenon as \emph{cross-modal transfer robustness}, namely, the ability of cross-modal geometry preservation to induce robust continual adaptation in the visual branch even when it is evaluated in a unimodal manner. As shown in Fig.~\ref{fig:V.}, the visual-branch variant (SeGP V.) consistently outperforms naive fine-tuning and the visual branch of MG-CLIP (MG-CLIP V.) \cite{huang2025mind}. More importantly, it achieves an exceptionally low forgetting rate, even lower than a frozen CLIP visual-classifier baseline (Continual V.). These results indicate that the gains brought by SeGP are not confined to cross-modal matching, but can be effectively distilled into a stronger and more stable visual representation stream, providing empirical support for our dual-path prediction that fuses cross-modal logits with prototype-based visual cues.

\subsection{Plasticity Analysis}
To further examine whether the proposed anchor-guided constraint over-regularizes new-class learning, we track the CE loss on raw current-task samples during training. We compare three settings: standard CE training, CE+ACGD using adversarial anchors, and CE+CGD directly imposed on raw new-task samples. Importantly, all curves in Fig.~\ref{fig:plasticity_ce_loss} report the supervised CE loss evaluated on raw current-task samples, rather than on the generated anchors.

As shown in Fig.~\ref{fig:plasticity_ce_loss}, CE+ACGD follows a similar decreasing trend to CE-only training on both CIFAR100 and ImageNet-R, indicating that the supervised optimization signal for raw new-task samples is not globally blocked by ACGD. In contrast, directly applying CGD on raw new-task samples leads to a noticeably higher CE loss and slower optimization. This comparison confirms that old-class distillation can suppress plasticity when it is imposed directly on incoming data, while our anchor-based ACGD avoids this direct sample-level conflict. Thus, our method improves the stability-plasticity trade-off while largely preserving the plasticity needed for effective new-class acquisition.

\subsection{Computational Cost}
Table~\ref{tab:lora_rank_inject} summarizes the computational overhead of our method in terms of trainable parameters and runtime. By fine-tuning CLIP with LoRA, our approach achieves stable continual adaptation with substantially fewer trainable parameters, resulting in a markedly smaller footprint than MoE-Adapter~\cite{yu2024boosting} and ZSCL~\cite{zheng2023preventing}, and also remaining more lightweight than CLAP~\cite{jha2024clap4clip} and DMNSP~\cite{kang2025dynamic}. Compared with standard fine-tuning baselines, the proposed distillation introduces only a minor training-time overhead: the overall training time increases by less than 20\%, while still being noticeably faster than MoE-Adapter~\cite{yu2024boosting} and ZSCL~\cite{zheng2023preventing}, highlighting the plug-and-play nature of our framework. At inference time, our method incurs a negligible additional cost beyond the original CLIP backbone (17.56~GFLOPs), since the auxiliary visual branch contributes only 0.00013~GFLOPs.

\subsection{Visualization}
\label{subsec:visualization}

\begin{figure}[t]
  \centering
  \includegraphics[width=\linewidth]{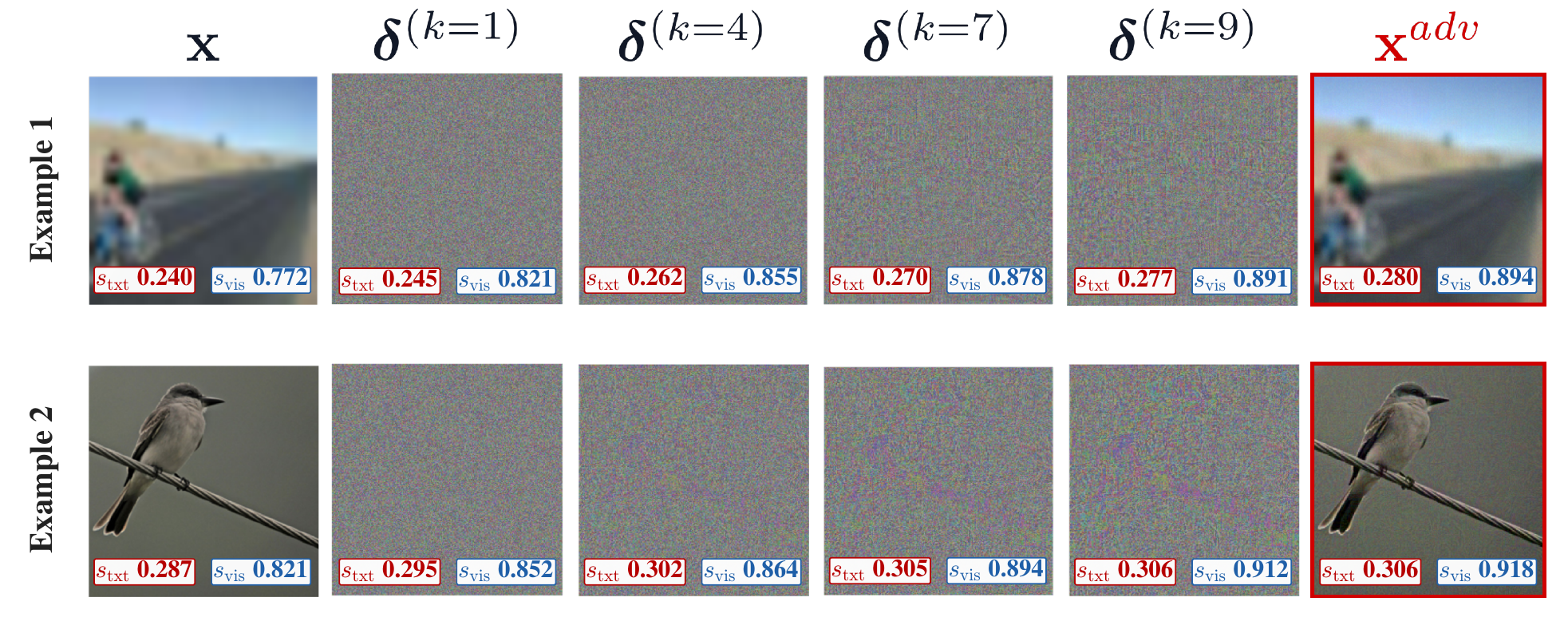}
  \vspace{-2em}
  \caption{DPGD optimization trajectories of adversarial anchors in visual-textual (CLIP) space and raw visual space.}
  \label{fig:vis-adv.}
\end{figure}

As shown in Fig.~\ref{fig:vis-adv.}, we visualize representative DPGD trajectories starting from seed samples. The middle columns show the perturbations \(\boldsymbol{\delta}^{(k)}\) at selected iterations, while the last column shows the final adversarial anchor \(\mathbf{x}^{adv}\). Although the perturbations gradually become structured, the final anchors remain visually close to the original seeds. Meanwhile, the annotated similarities show that both the target-text similarity \(s_{\mathrm{txt}}\) and the raw visual-prototype similarity \(s_{\mathrm{vis}}\) increase along the DPGD process. This indicates that DPGD moves the selected seeds toward the target old-class semantic neighborhood in both the CLIP visual-text space and the raw visual space.

\begin{figure}
    \centering
    \includegraphics[width=\linewidth]{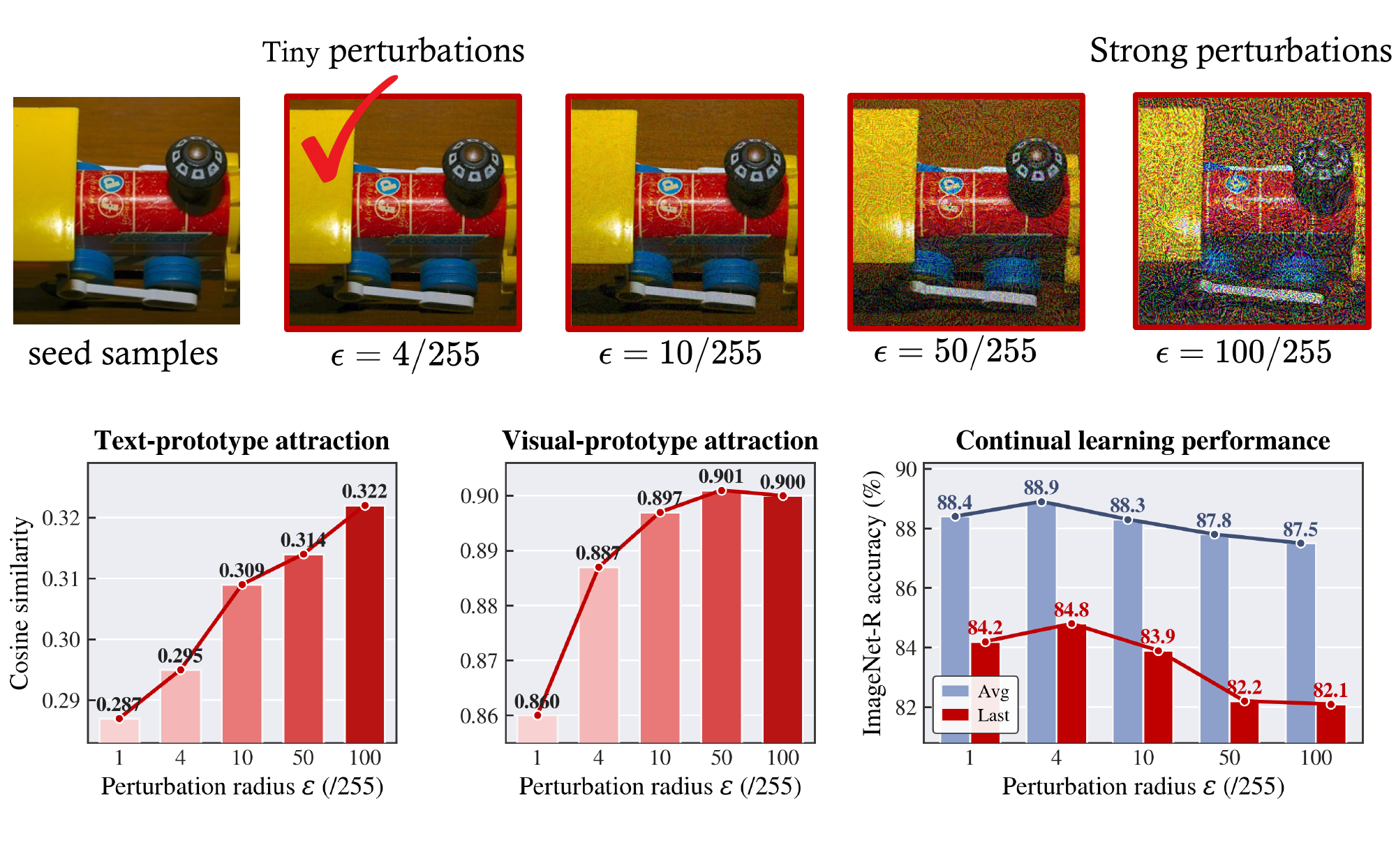}
    \vspace{-2.5em}
    \caption{Effect of perturbation budget on adversarial anchors. We visualize anchors generated with increasing $\ell_\infty$ radius $\epsilon$, and report their similarities to the targeted old text embedding and raw visual prototype, together with the resulting ImageNet-R CL performance.}
    \label{fig:adv-radius}
\end{figure}

In addition, Fig.~\ref{fig:adv-radius} visualizes the effects of different perturbation budgets. For a fair comparison, we keep \(K_{\mathrm{adv}}\) fixed and adjust the DPGD step size proportionally with the perturbation radius, i.e., \(\gamma_{\epsilon}\approx\epsilon/K_{\mathrm{adv}}\). Larger perturbation radii can produce anchors with higher similarity to the targeted old-class text embedding and raw visual prototype. However, distillation on these large-radius anchors yields worse continual learning performance than the tiny-perturbation setting (e.g., \(\epsilon=4/255\)). This suggests that overly large perturbations may drive anchors away from the original natural-image visual domain and introduce noisy distillation signals. Therefore, effective adversarial anchors should remain local probes of vulnerable semantic-interface neighborhoods, rather than unrestricted perturbations.

To directly inspect how our method mitigates semantic drift around the old-new semantic interface, we further visualize boundary old-class samples in a semantic-coordinate plane on CIFAR100. For an old-class sample \((\mathbf{x},y)\) with \(y\in\mathcal{C}_{<t}\), the horizontal coordinate \(d_{\mathrm{old}}\) denotes its distance to the ground-truth old-class text axis, computed as \(1-(\bar{\mathbf{v}}(\mathbf{x}))^\top\mathbf{u}_{y}\). The vertical coordinate \(d_{\mathrm{new}}^{\mathrm{near}}\) denotes its distance to the nearest new-class text axis, computed as \(1-\max_{c\in\mathcal{C}_{t}}(\bar{\mathbf{v}}(\mathbf{x}))^\top\mathbf{u}_{c}\). Here, \(\bar{\mathbf{v}}(\mathbf{x})\) and \(\mathbf{u}_{c}\) are evaluated using the corresponding model state before or after task learning when plotting the drift arrows. Thus, the \textit{lower-right direction} indicates harmful semantic drift, where old visual patterns move away from their original old-class semantics and become closer to newly introduced class semantics. This diagnostic visualization uses old samples only for analysis and does not affect the exemplar-free training protocol.

\begin{figure}[t]
    \centering
    \includegraphics[width=0.85\linewidth]{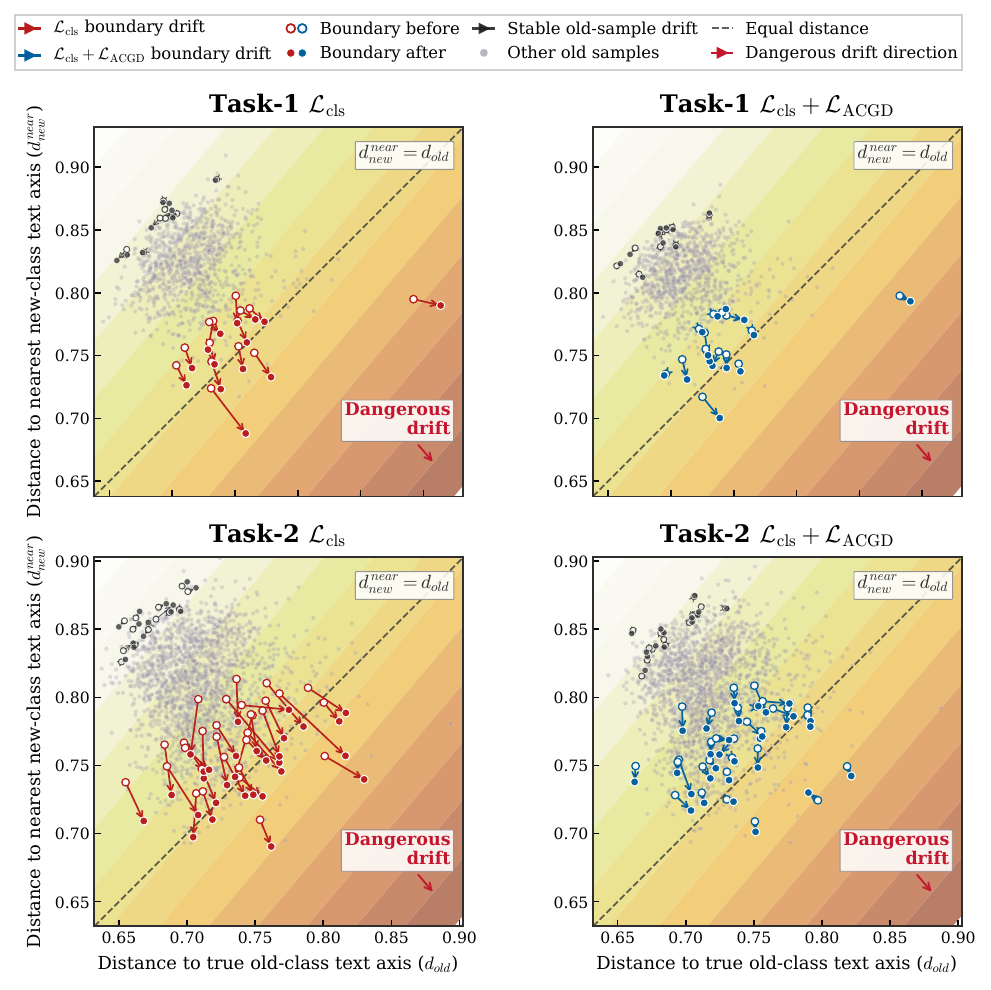}
    \vspace{-1em}
    \caption{Visualization of boundary semantic drift after task learning on CIFAR-100.}
    \label{fig:boundary_semantic_drift}
\end{figure}

As shown in Fig.~\ref{fig:boundary_semantic_drift}, CE-only training($\mathcal{L}_{\text{cls}}$) produces clear drift vectors toward the lower-right direction, especially for boundary samples after Task-2, suggesting that ambiguous old-new shared visual patterns are aggressively re-explained by new-class text semantics. In contrast, with ACGD, the drift vectors are shorter and less aligned with this harmful direction, and boundary samples remain closer to the equal-distance region between old and new semantics. Meanwhile, stable old samples remain concentrated in the safe region. These observations further support that harmful semantic drift is concentrated around vulnerable boundary neighborhoods, and that SeGP-CL can effectively suppress such drift through targeted anchor-guided geometry preservation.

\subsection{Evaluation Under Task-Incremental Protocols}
\label{subsec:taskil_mtil}

To further examine the generality of SeGP-CL beyond the class-incremental protocol, we also evaluate it under task-incremental learning (TIL) settings.

\textit{1) Conventional TIL:} We further evaluate SeGP-CL under task-incremental protocols. In both class-incremental learning (CIL) and TIL, different tasks introduce disjoint semantic classes, i.e., \(\mathcal{C}_i\cap\mathcal{C}_j=\emptyset\). The difference lies in the inference label space: CIL predicts over all seen classes \(\mathcal{C}_{\le t}\), while TIL predicts within the task-specific label set \(\mathcal{C}_{t}\). Therefore, standard TIL provides a complementary setting to examine whether semantic-geometry preservation remains effective when task identity is available at inference. As shown in Table~\ref{tab:taskil_results}, our method achieves the best \textit{Avg} and \textit{Last} accuracy across both datasets and task splits. On CIFAR100, our method improves the strongest competing \textit{Last} accuracy from 94.5 to 95.7 under the 5-task split and from 94.8 to 95.9 under the 10-task split. On ImageNet-R, it improves \textit{Last} accuracy from 91.9 to 93.5 and from 94.2 to 95.2, respectively. These results show that constraining semantic geometry still brings clear benefits under the TIL protocol.

\begin{table}[t]
\centering
\caption{Results under the standard TIL setting. The task identity is given at inference, and prediction is performed within the corresponding task label space. We report \textit{Avg} and \textit{Last} accuracy under 5-task and 10-task splits.}
\label{tab:taskil_results}
\setlength{\tabcolsep}{7pt}
\renewcommand{\arraystretch}{1.08}
\resizebox{\columnwidth}{!}{%
\begin{tabular}{l cc cc cc cc}
\toprule[1.pt]
\multirow{3}{*}{\textbf{Method}} &
\multicolumn{4}{c}{\textbf{CIFAR100}} &
\multicolumn{4}{c}{\textbf{ImageNet-R}} \\
\cmidrule(lr){2-5}\cmidrule(lr){6-9}
&
\multicolumn{2}{c}{\textbf{5 Tasks}} &
\multicolumn{2}{c}{\textbf{10 Tasks}} &
\multicolumn{2}{c}{\textbf{5 Tasks}} &
\multicolumn{2}{c}{\textbf{10 Tasks}} \\
\cmidrule(lr){2-3}\cmidrule(lr){4-5}
\cmidrule(lr){6-7}\cmidrule(lr){8-9}
&
\textit{Avg} & \textit{Last} &
\textit{Avg} & \textit{Last} &
\textit{Avg} & \textit{Last} &
\textit{Avg} & \textit{Last} \\
\midrule
Continual-CLIP~\cite{thengane2022clip} &
82.5 & 81.8 & 87.9 & 86.9 &
89.3 & 88.9 & 84.5 & 84.4 \\

RAPF~\cite{huang2025class} &
92.5 & 92.2 & 95.2 & 94.8 &
91.0 & 91.2 & 94.1 & 93.9 \\

DesCLIP~\cite{he2025desclip} &
89.3 & 89.1 & 93.0 & 92.5 &
90.4 & 90.5 & 93.5 & 93.1 \\

ENGINE~\cite{zhou2025external} &
90.2 & 90.0 & 93.8 & 93.1 &
89.8 & 88.3 & 93.1 & 91.4 \\

MG-CLIP~\cite{huang2025mind} &
94.9 & 94.5 & 95.0 & 94.7 &
92.1 & 91.9 & 94.1 & 94.2 \\

\rowcolor[HTML]{D7D2F5}
Ours &
\textbf{96.0} & \textbf{95.7} &
\textbf{96.2} & \textbf{95.9} &
\textbf{93.3} & \textbf{93.5} &
\textbf{94.9} & \textbf{95.2} \\

\bottomrule[1.pt]
\end{tabular}
}
\end{table}

\begin{table}[t]
\centering
\caption{Comparison on the MTIL benchmark under Order-I and Order-II~\cite{zheng2023preventing}.}
\label{tab:mtil_results}
\setlength{\tabcolsep}{7.0pt}
\renewcommand{\arraystretch}{1.08}
\resizebox{\columnwidth}{!}{%
\begin{tabular}{lc ccc ccc}
\toprule[1.pt]
\multirow{2}{*}{\textbf{Method}} &
\multirow{2}{*}{\makecell{\textbf{Aux.}\\\textbf{Source}}} &
\multicolumn{3}{c}{\textbf{Order-I}} &
\multicolumn{3}{c}{\textbf{Order-II}} \\
\cmidrule(lr){3-5}\cmidrule(lr){6-8}
&
&
\textit{Transfer} &
\textit{Avg} &
\textit{Last} &
\textit{Transfer} &
\textit{Avg} &
\textit{Last} \\
\midrule
\multicolumn{8}{l}{\textit{Methods without auxiliary data}} \\

Zero-shot &
None &
69.4 & 65.3 & 65.3 &
65.4 & 65.3 & 65.3 \\

Finetune &
None &
44.6 & 55.9 & 77.3 &
46.6 & 56.2 & 67.4 \\

\rowcolor[HTML]{EEEEEE}
Finetune + \(l_2\)~\cite{zheng2023preventing} &
None &
61.0 & 62.7 & 75.9 &
60.6 & 68.8 & 77.2 \\

LwF~\cite{li2017learning} &
None &
56.9 & 64.7 & 74.6 &
53.2 & 62.2 & 71.9 \\

LwF-VR~\cite{ding2022dont} &
None &
57.2 & 65.1 & 76.6 &
53.1 & 60.6 & 68.3 \\

WiSE-FT~\cite{wortsman2022robust} &
None &
52.3 & 60.7 & 77.7 &
51.0 & 61.5 & 72.2 \\

\rowcolor[HTML]{D7D2F5}
Ours-\textit{OnlyCLIP} &
None &
64.7 & \textbf{68.5} & \textbf{80.5} &
61.9 & \textbf{72.5} & \textbf{80.9} \\

\midrule
\multicolumn{8}{l}{\textit{Methods with external auxiliary data}} \\

ZSCL~\cite{zheng2023preventing} &
Ext. Ref. &
68.1 & 75.4 & 83.6 &
64.2 & 74.5 & 83.4 \\

GIFT~\cite{wu2025synthetic} &
Gen. Img. &
69.3 & 77.3 & 86.0 &
65.9 & 75.7 & 85.3 \\

\bottomrule[1.pt]
\end{tabular}
}
\end{table}

\textit{2) Multi-domain TIL:} We also evaluate SeGP-CL under the broader MTIL benchmark following recent VLM-based studies~\cite{zheng2023preventing,wu2025synthetic}. MTIL is a cross-domain TIL protocol where each task corresponds to one dataset/domain and the task identity is available at inference. We follow the dataset composition, Order-I/Order-II task sequences, and the Finetune+\(l_2\) baseline setting from the original benchmark~\cite{zheng2023preventing}. Following this protocol, \textit{Transfer} measures the average zero-shot performance on future unseen tasks before they are learned, \textit{Avg} denotes the average accuracy over all tasks and training stages, and \textit{Last} denotes the final average accuracy after all tasks are learned. As shown in Table~\ref{tab:mtil_results}, our \textit{OnlyCLIP} variant consistently improves over the Finetune+\(l_2\) baseline under both MTIL orders without using external auxiliary data. It achieves the best \textit{Avg} and \textit{Last} scores among methods without auxiliary data, showing that semantic-geometry preservation also benefits multi-domain task-incremental learning. Although ZSCL \cite{zheng2023preventing} and GIFT \cite{wu2025synthetic} report higher absolute scores, they rely on external reference data or diffusion-generated images. In contrast, our \textit{OnlyCLIP} variant keeps a stricter data assumption and isolates the contribution of semantic-geometry preservation itself.

\section{Limitations}
Although SeGP-CL is exemplar-free and does not store any real images from previous tasks, it maintains a lightweight historical memory of raw texts and visual prototypes. This introduces modest storage and computational overhead for anchor construction and geometry regularization.

Similar to recent VLM-based CL frameworks such as RAPF~\cite{huang2025class}, ENGINE~\cite{zhou2025external}, and MG-CLIP~\cite{huang2025mind}, SeGP-CL is mainly designed for continual adaptation scenarios where downstream tasks progressively introduce new visual and semantic concepts. Thus, the core focus is expanding semantic knowledge while preserving previously established vision-language alignment. Other practical scenarios may follow different assumptions. For example, strict shared-label Domain-IL keeps the label space unchanged while the input distribution shifts across domains, and usually requires domain identification or domain-conditioned adaptation. Since SeGP-CL does not explicitly model domain identity, it may not fully address such shared-label domain-incremental scenarios.

Finally, the effectiveness of the semantic-relevance structure depends on the quality of prompt templates, and the current design may be limited under severely out-of-distribution scenarios where vision-language relations deviate substantially from the pretrained alignment.

\section{Conclusions}
Current continual learning methods for vision-language models (VLMs) often update the visual encoder under new-task supervision without explicitly protecting the cross-modal semantic geometry inherited from pretraining or previous tasks. Empirically, such unprotected updates are often reflected by an amplified geometric drift around the old-new cross-modal semantic interface, i.e., in the vulnerable neighborhoods where shared visual patterns are most susceptible to being re-interpreted by new-task texts, which accelerates forgetting. In this work, we propose SeGP-CL, a concise framework for semantic geometry preservation during continual learning. Specifically, we construct a compact set of adversarial anchors via DPGD to probe drift-prone neighborhoods, and introduce anchor-guided cross-modal geometry distillation together with a lightweight text semantic-geometry regularization to stabilize the textual reference frame during adaptation. In post-training stage, we further perform anchor-induced prototype transfer and adopt a dual-path inference scheme that fuses cross-modal and prototype cues for robust prediction. Extensive experiments validate the effectiveness of SeGP-CL, consistently accumulating downstream-task knowledge and reducing forgetting compared with existing VLM-based continual learning baselines.

\bibliographystyle{IEEEtran}

\bibliography{SeGP_ref}

\begin{IEEEbiography}[{\includegraphics[width=1in,height=1.25in,clip,keepaspectratio]{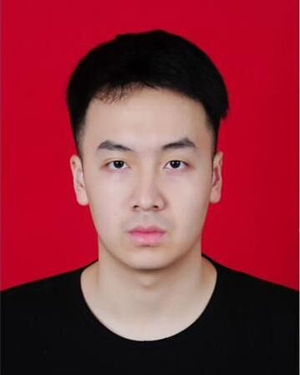}}]{Chiyuan He} recieved his M.S. degree in Information and Communication Engineering at the University of Electronic Science and Technology of China (UESTC). He is currently pursuing the Ph.D. degree in UESTC. His main research interests include lifelong learning and multimodal intelligence.
\end{IEEEbiography}

\begin{IEEEbiography}
[{\includegraphics[width=1in,height=1.25in,clip,keepaspectratio]{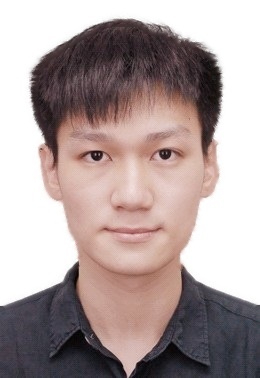}}]{Zihuan Qiu} is currently pursuing the Ph.D. degree in University of Electronic Science and Technology of China, Chengdu, China. His current research interests include continual learning and machine learning.
\end{IEEEbiography}

\begin{IEEEbiography}[{\includegraphics[width=1in,height=1.25in,clip,keepaspectratio]{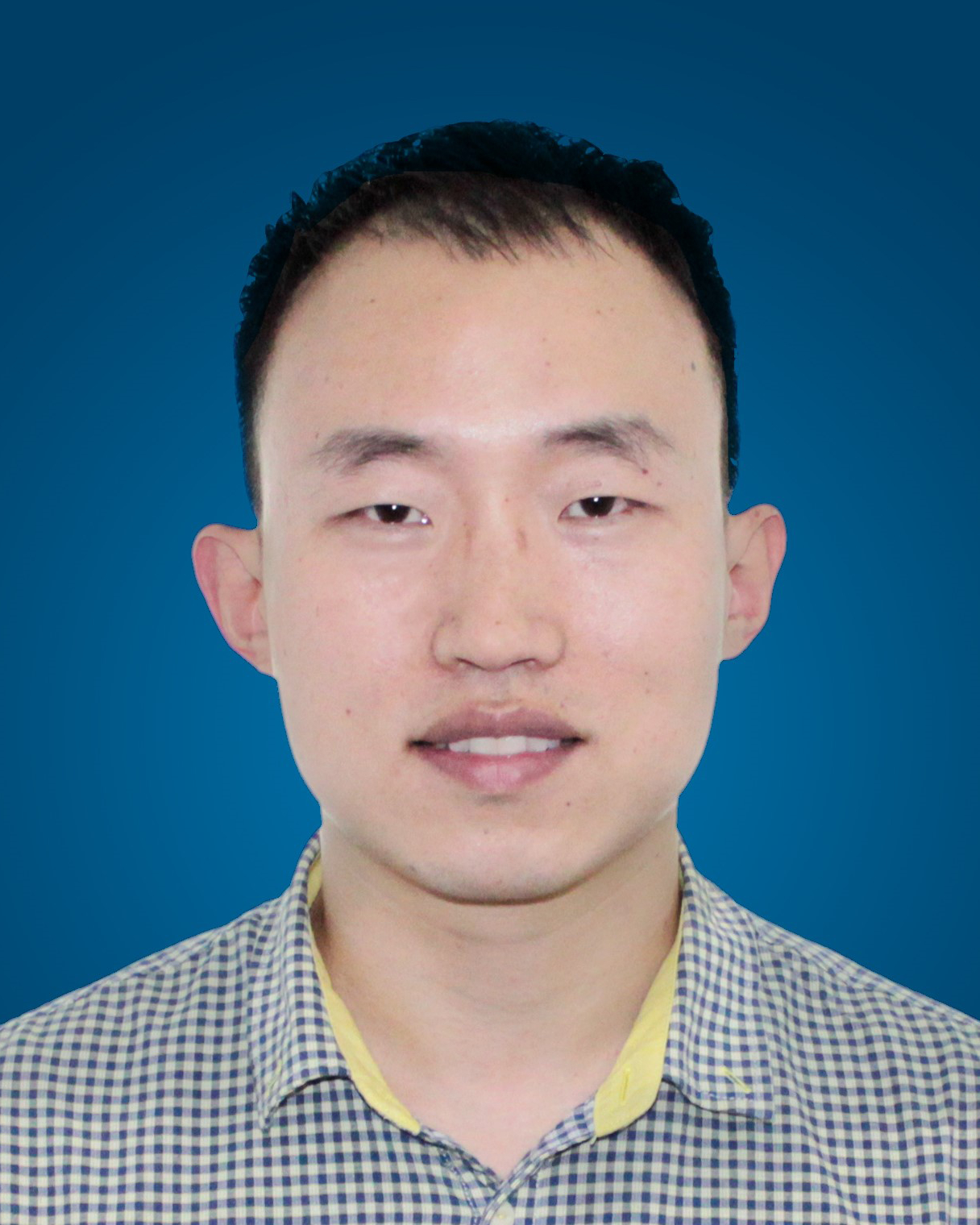}}]{Fanman Meng} (Member, IEEE) received the Ph.D. degree in Signal and Information Processing from the University of Electronic Science and Technology of China, Chengdu, China, in 2014. From 2013 to 2014, he was a Research Assistant with the Division of Visual and Interactive Computing, Nanyang Technological University, Singapore. He is currently a Professor with the School of Information and Communication Engineering, University of Electronic Science and Technology of China. He has authored or co-authored numerous technical articles in well-known international journals and conferences. His current research interests include image segmentation and object detection. 
\end{IEEEbiography}

\begin{IEEEbiography}
[{\includegraphics[width=1in,height=1.25in,clip,keepaspectratio]{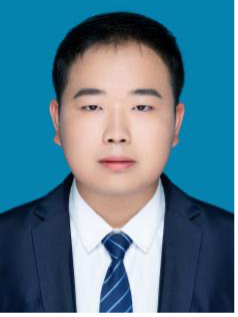}}]{Runtong Zhang} is currently working toward the Ph.D. degree with the School of Information and Communication Engineering, University of Electronic Science and Technology of China, Chengdu, 611731, China. His current research interests include semantic segmentation, domain generalization, few/zero-shot learning.
\end{IEEEbiography}

\begin{IEEEbiography}
[{\includegraphics[width=1in,height=1.25in,clip,keepaspectratio]{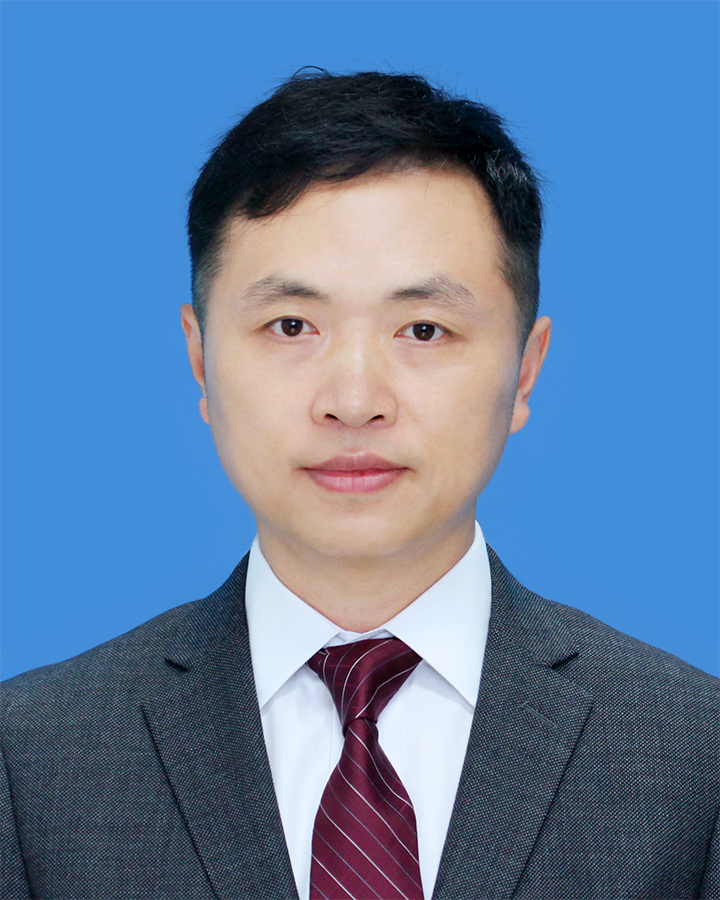}}]{Linfeng Xu} (Member, IEEE) received the Ph.D. degree in Signal and Information Processing from the School of Electronic Engineering, University of Electronic Science and Technology of China (UESTC), Chengdu, China, in 2014. From December 2014 to December 2015, he was with the Ubiquitous Multimedia Laboratory, the State University of New York at Buffalo, USA, as a visiting scholar. He is currently an Associate Professor with the School of Information and Communication Engineering, UESTC. His research interests include machine learning, computer vision, visual signal processing, artificial intelligence theory and applications. He served as a Local Arrangement Chair for ISPACS 2010 and VCIP 2016.
\end{IEEEbiography}

\begin{IEEEbiography}[{\includegraphics[width=1in,height=1.25in,clip,keepaspectratio]{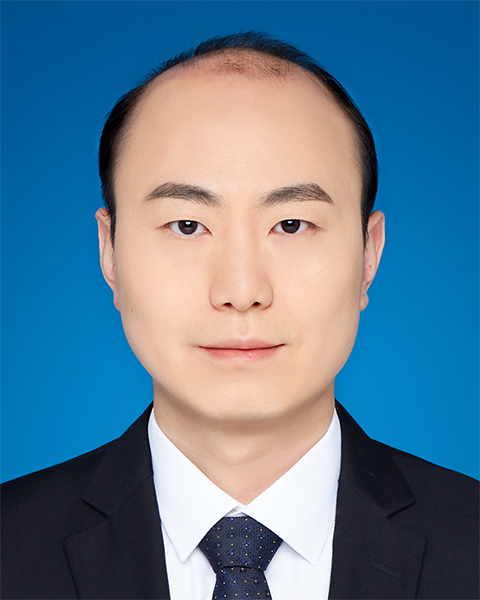}}]{Qingbo Wu} (Member, IEEE) received the Ph.D. degree in signal and information processing from the University of Electronic Science and Technology of China in 2015. From February 2014 to May 2014, he was a Research Assistant with the Image and Video Processing (IVP) Laboratory, Chinese University of Hong Kong. From October 2014 to October 2015, he served as a Visiting Scholar with the Image and Vision Computing (IVC) Laboratory, University of Waterloo. He is currently a Professor with the School of Information and Communication Engineering, University of Electronic Science and Technology of China. His research interests include image/video coding, quality evaluation, perceptual modeling and processing. He has served as Area Chair for ACM MM 2024-2025, VCIP 2016, Session Chair for ACM MM 2021, ICMCT 2022, TPC/PC member of AAAI 2021-2023, APSIPA ASC 2020-2021, CICAI 2021-2023. He was also a Guest Editor of Remote Sensing and Frontiers in Neuroscience. 
\end{IEEEbiography}

\begin{IEEEbiography}[{\includegraphics[width=1in,height=1.25in,clip,keepaspectratio]{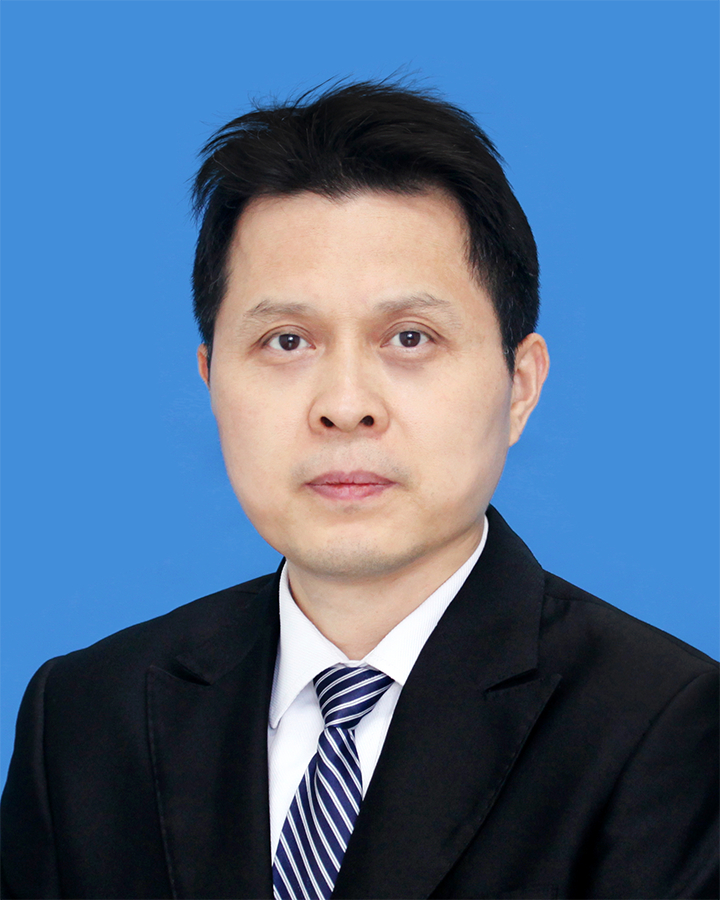}}]{Hongliang Li} (Senior Member, IEEE) received his Ph.D. degree in Electronics and Information Engineering from Xi’an Jiaotong University, China, in 2005. From 2005 to 2006, he joined the visual signal processing and communication laboratory (VSPC) of the Chinese University of Hong Kong (CUHK) as a Research Associate. From 2006 to 2008, he was a Postdoctoral Fellow at the same laboratory in CUHK. He is currently a Professor in the School of Information and Communication Engineering, University of Electronic Science and Technology of China. His research interests include image segmentation, object detection, image and video coding, visual attention, and multimedia processing. 

Dr. Li has authored or co-authored numerous technical articles in well-known international journals and conferences. He is a co-editor of a Springer book titled “Video segmentation and its applications”. Dr. Li is involved in many professional activities. He received the 2019 and 2020 Best Associate Editor Awards for IEEE Transactions on Circuits and Systems for Video Technology (TCSVT), and the 2021 Best Editor Award for Journal on Visual Communication and Image Representation. He served as a Technical Program Chair for VCIP 2016 and PCM 2017, General Chairs for ISPACS 2017 and ISPACS 2010, a Publicity Chair for IEEE VCIP 2013, a Local Chair for the IEEE ICME 2014, Area Chairs for VCIP 2022 and 2021, and a Reviewer committee member for IEEE ISCAS from 2018 to 2022. He served as an Associate Editor of IEEE Transactions on Circuits and Systems for Video Technology (2018-2021). He is now an Associate Editor of Journal on Visual Communication and Image Representation, IEEE Open Journal of Circuits and Systems, and an Area Editor of Signal Processing: Image Communication (Elsevier Science). He is selected as the IEEE Circuits and Systems Society Distinguished Lecturer for 2022-2023.
\end{IEEEbiography}

\end{document}